\documentclass[journal]{IEEEtran}
\usepackage{xcolor}

\usepackage{enumitem}
\usepackage{amsmath}
\usepackage{amssymb}
\usepackage{subfigure}
\usepackage{color}
\usepackage{graphicx}
\usepackage{algorithm,algpseudocode,algorithmicx}
\usepackage{amsmath,latexsym,amssymb,amsthm,array,amsfonts,algorithm,algpseudocode,booktabs,graphicx,subfigure,multirow,cuted,stfloats}
\usepackage{footmisc}
\usepackage{cite}
\usepackage{units}
\usepackage{tabularx}
\usepackage{multirow}
\usepackage{bm}
\usepackage{float}
\usepackage{soul}
\newlength\savewidth

\ifCLASSINFOpdf
\else
\fi

\begin{document}
\title{Real-time Federated Evolutionary Neural Architecture Search}

\author{Hangyu~Zhu and
        Yaochu~Jin, \emph{Fellow}, \emph{IEEE}
\thanks{Hangyu Zhu and Yaochu Jin are with the Department of Computer Science, University of Surrey, Guildford, GU2 7XH, United Kingdom. Email: \{hangyu.zhu;yaochu.jin\}@surrey.ac.uk. (\textit{Corresponding author: Yaochu Jin})}
}

\maketitle

\begin{abstract}
Federated learning is a distributed machine learning approach to privacy preservation and two major technical challenges prevent a wider application of federated learning. One is that federated learning raises high demands on communication, since a large number of model parameters must be transmitted between the server and the clients. The other challenge is that training large machine learning models such as deep neural networks in federated learning requires a large amount of computational resources, which may be unrealistic for edge devices such as mobile phones. The problem becomes worse when deep neural architecture search is to be carried out in federated learning. To address the above challenges, we propose an evolutionary approach to real-time federated neural architecture search that not only optimizes the model performance but also reduces the local payload. During the search, a double-sampling technique is introduced, in which for each individual, a randomly sampled sub-model of a master model is transmitted to a number of randomly sampled clients for training without reinitialization. This way, we effectively reduce computational and communication costs required for evolutionary optimization and avoid big performance fluctuations of the local models, making the proposed framework well suited for real-time federated neural architecture search.
\end{abstract}

\begin{IEEEkeywords}
Federated learning, neural architecture search, multi-objective evolutionary optimization, real-time optimization, deep neural networks, communication cost
\end{IEEEkeywords}

\IEEEpeerreviewmaketitle

\section{Introduction}
\IEEEPARstart{S}{tandard} centralized machine learning methods require to collect training data from distributed users and stored them on a single server, which suffers from a high risk of leaking users' private information. Therefore, a distributed approach called federated learning \cite{mcmahan2016communication} was proposed to preserve data privacy, enabling multiple local devices to collaboratively train a shared global model while the training data remain to be deployed on the edge devices. Consequently, the central server has no access to the private raw data and the client privacy is protected.

However, federated learning demands a large amount of communication resources in contrast to the conventional centralized learning paradigm, since updating the global model needs to frequently download and upload model parameters between the server and edge clients. To mitigate this problem, a large body of research work has been carried out to reduce the communication costs in federated learning. The most popular approaches include compression and sub-sampling of the client uploads \cite{shokri2015privacy, konevcny2016federated, caldas2018expanding} or quantization of the weights of the models \cite{han2015deep}. Most recently, Chen \emph{et al.} \cite{chen2019communication} suggest a layer-wise parameter update algorithm to reduce the number of parameters to be transmitted between the server and the clients. In addition, Zhu \emph{et al.} \cite{zhu2019multi} uses a multi-objective evolutionary algorithm (MOEA) to simultaneously enhance the model performance and communication efficiency.

Little work has been reported on offline optimization of the architecture of deep neural networks (DNNs) for federated learning, let alone real-time neural architecture search (NAS) suited for the federated environment. In the field of centralized machine learning, Zoph \emph{et al.} \cite{zoph2016neural} present some early work on NAS using reinforcement learning, which, however, consumes a plenty of computational resources. To mitigate this problem, Pham \emph{et al.} \cite{pham2018efficient} introduce a directed acyclic graph (DAG) based neural architecture representation to significantly accelerate the search speed without much degradation of the learning performance by using weight sharing technique.

Recently, evolutionary approaches to NAS have received increasing attention  \cite{suganuma2017genetic,real2019regularized,real2017large,liu2017hierarchical}, and hybrid methods that combine evolutionary search with the gradient method \cite{liu2018darts} have also been reported \cite{dong2019searching}. To reduce the computational cost, surrogate ensembles have been introduced in evolutionary NAS, which has been shown promising in reducing the computational complexity without deteriorating the learning performance \cite{Sun2019}.

Most existing search strategies for NAS in a centralized learning environment are not well suited for federated NAS for the following reasons. First, most current search strategies for centralized learning focus on improving the model performance without paying much attention to the model size and computation cost. Client devices like mobile phones cannot afford computationally intensive model training and bandwidth restrictions do not allow very large models to be transmitted frequently between the server and clients. Second, many NAS techniques for accelerating model training \cite{pham2018efficient,real2019regularized,liu2017hierarchical,dong2019searching,
liu2018darts,tan2019mnasnet} adopt transfer learning techniques \cite{pan2009survey,torrey2010transfer} by searching upon cell based small models and transferring the found promising cell structures to large models. Such techniques are not directly applicable to federated learning, because such cell transfer methods may cause model divergence in a distributed training scheme and learning transferred models from scratch will consume extra communication resources. Finally, modern deep neural networks may fail to work in federated learning as learnable parameters in the batch normalization layer \cite{ioffe2015batch} may degrade the global model performance after model aggregation, because the locally trained models may have very different weights in mean and variance.

Note that our previous work on multi-objective evolutionary federated optimization \cite{zhu2019multi} is an offline evolutionary approach to NAS, like most conventional evolutionary optimization algorithms. In offline evolutionary NAS, the parameters of a newly generated offspring model are randomly reinitialized and trained from scratch before the model is evaluated on a validation dataset, requiring a large amount of  computational resources. What is worse, the performance of the reinitialized models will dramatically degrade, making the offline optimization approach infeasible for realtime application of federated learning, such as online recommendation systems \cite{ammad2019federated}.

Therefore, offline federated evolutionary optimization of neural networks is not applicable to the real-world applications and it is highly desirable to develop a framework for real-time federated evolutionary neural architecture search. The main contributions of this work are summarized as follows:
\begin{enumerate}
\item A double-sampling technique is proposed that randomly samples a sub-network of the global model, whose parameters are transmitted to a few randomly sampled local devices without replacement. The number of devices to be sampled depends on the ratio between the number of connected local clients and the number of individuals in the population. The double-sampling technique brings about two advantages. First, only a sub-network needs to be trained on local devices, significantly reducing the number of parameters to be uploaded from the local devices to the server. Second, sampling of the clients without replacement makes sure that each local device needs to train only one sub-network for once at each generation. The above two features together make the proposed real-time evolutionary method substantially different from offline evolutionary NAS, where the whole neural network must be trained on all local devices for fitness evaluations, and each device needs to train all networks in the population. To the best of our knowledge, it is the first time that a real-time evolutionary NAS algorithm has been developed for the federated learning framework.

\item An aggregation strategy is developed that updates the global model based on the sub-networks sampled and trained at each generation. Thus, in the proposed real-time evolutionary NAS, each generation is equivalent to a training round in the traditional federated learning. In addition, the weights of the sampled sub-networks are inherited before it is trained on a local device, accelerating the convergence and avoiding dramatic performance deterioration caused by random reinitialization.
\end{enumerate}

Extensive comparative studies are performed to verify the performance of the proposed real-time federated evolutionary NAS by comparing the learning performance and computational complexity of the models it obtains with that of the standard ResNet18 \cite{he2016deep} on both \emph{IID} and \emph{non-IID} data. We also show the proposed real-time evolutionary NAS is at least five times faster than the offline evolutionary NAS method.

\section{Background}
In this section, a review of federated learning is given at first, followed by an introduction to NAS for deep neural networks. Then, the basic of a multi-objective evolutionary algorithms will be introduced. Finally, a brief discussion is given to clarify the differences between the offline and real-time evolutionary optimization frameworks.

\subsection{Federated Learning}
As mentioned before, federated learning is an emerging decentralized privacy-preserving model training technology that enables local users to training a global model without uploading their private local data to a central server. A conventional federated learning algorithm called federated averaging (FedAvg) algorithm is shown in \textbf{Algorithm \ref{fedavg}}. In the following,  we briefly introduce this algorithm.

\begin{algorithm}[htbp]\footnotesize{
\caption{FederatedAveraging. $K$ indicates the total number of clients; $B$ is size of mini-batches, $ E\ $ is equal to the number of training iterations, $ \eta \ $ is the learning rate, $n$ is the total number of data pairs on all distributed clients, and $n_{k}$ is the number of data pairs on client $k$.}
\algblock{Begin}{End}
\label{fedavg}
\begin{algorithmic}[1]
\State \textbf{Server Update: }
\State Initialize $ {\theta(0)}\ $
\State Let $\theta(t) = \theta(0)$
\For {each communication round $ t = 1,2,...\ $}
\State Select $ m = C \times K $ clients, $ C \in (0,1)\ $ clients
\State Download $ {\theta(t)}\ $ to each client $k$
\For {each client $ k \in m\ $}
\State Wait \textbf{Client $k$ Update} for synchronization
\State $ {\theta(t)} = \sum\limits_{k = 1}^m {\frac{{{n_k}}}{n}{\theta(t) ^k}} \ $
\EndFor
\EndFor \\
\State \textbf{Client $k$ Update: }
\State $ {\theta ^k} = {\theta(t)}\ $
\For {each iteration from 1 to $E$}
\For {batch $ b \in B\ $}
\State $ {\theta ^k} = {\theta ^k} - \eta \nabla {L_k}({\theta ^k},b)\ $
\EndFor
\EndFor
\State return ${\theta ^k}$ to server
\end{algorithmic}}
\end{algorithm}

\subsubsection{Server Side}
The model parameters $\theta(0)$ are initialized once at the beginning of FedAvg algorithm, which is then sent to all selected $m = C \times K$ clients, where $K$ is the number of total clients and $C$ is the fraction of all clients between $0$ and $1$. After all $m$ clients update and send the updated local model parameters $\theta ^{k}$ back to the central server, the parameters $\theta(t)$ of the global server model will be replaced by the weighted averaging of each client's model parameters $\theta ^{k}$.

\subsubsection{Client Side}
The local model parameters $\theta ^k$ are replaced by the downloaded global model parameters $\theta_{t}$. Then the local model parameters are updated by the batch stochastic gradient descent (SGD) algorithm \cite{bottou1991stochastic}, where $b$ is the local learning batch size. After local training, the learned model parameters $\theta^{k}$ will be sent back to the central server for global model aggregation.

An alternative approach to local execution is to calculate and upload the local model gradients only. Then, they are aggregated on the server by $\theta(t)=\theta(t)-\sum_{k=1}^{m}\frac{n_{k}}{n}g^{k}$, where $g^{k}$ represents local gradients. This method is beneficial to reduce the local computation consumption while sharing the same computing result with the previous one.


\subsection{Neural Architecture Search}
Deep learning \cite{lecun2015deep} has been extremely successful in the field of image recognition and speech recognition. However, most deep neural network (DNN) models are manually designed by human experts, and not until very recently has increasing attention been attracted to automatically search for a good model architecture using NAS methods. The search space in NAS depends on the neural network model in use and in this work, we limit our discussions to convolutional neural networks (CNNs) \cite{lecun1995convolutional, krizhevsky2012imagenet}.

\begin{figure}
\centering
\includegraphics[height=2cm, width=6cm]{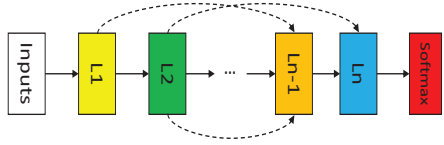}
\caption{An illustrative example of CNN represented in a macro search space.}
\label{macro}
\end{figure}

\begin{figure}
\centering
\includegraphics[height=5cm, width=7cm]{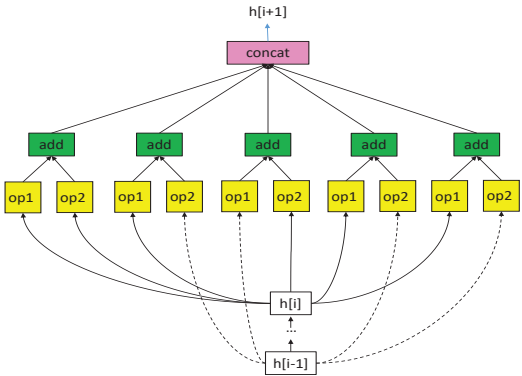}
\caption{A typical structure of the normal cell represented in a micro search space, where each block receives the outputs from the previous cell $h[i]$ and the cell before the previous cell $h[i-1]$ as its inputs, which are then connected to two operations, denoted by 'op' in the figure. Finally all the branches are concatenated at the output of this cell.}
\label{micro}
\end{figure}

Roughly speaking, the search space for NAS can be categorized into macro and micro search spaces \cite{pham2018efficient}. The macro search space is over the entire CNN model, as shown in Fig. \ref{macro}. The whole model consists of $n$ sequential layers where the dashed lines are shortcut connections, similar to ResNet \cite{he2016deep}. So the macro search space aims to represent a good structured model in terms of the number of hidden layers $n$, operations types (e.g. convolution), model hyper parameters (e.g. convolutional kernel size), and the link methods for shortcut connections. Different from the macro search space, micro search space only covers repeated motifs or cells \cite{szegedy2016rethinking, huang2017densely} in the whole model structure. And these cells are built in complex multi-branch operations \cite{szegedy2017inception, szegedy2015going} as shown in Fig. \ref{micro}, where the given structure contains two inputs $h[i]$ and $h[i-1]$ coming from two previous layers and only one concatenated output.

Much work has been done that uses reinforcement learning (RL) for NAS \cite{zoph2016neural,zoph2018learning,pham2018efficient,baker2016designing,zhong2018practical}.  RL-based search methods always adopt a recurrent neural network (RNN) \cite{schuster1997bidirectional} as a controller to sample a new candidate model to be trained and then use the model performance as the reward score. And then this score can be used to update the controller for sampling a better child model in the next iteration.

Apart from RL, evolutionary algorithms (EAs) based approaches are often used to deal with multi-objective optimization problems in NAS. Different from the previous neuro-evolution techniques \cite{angeline1994evolutionary,yao1999evolving} that aim to optimize both the weights and architecture of neural networks, EA based NAS only optimizes the model architecture itself, and the model parameters are trained using conventional gradient descent methods \cite{real2017large,suganuma2017genetic,liu2017hierarchical,real2019regularized}. Evolutionary NAS starts with randomly generating a population of parent models with different structures and train them in parallel. When all the models are learned for some pre-defined epochs, fitness values are calculated by evaluating the models on the validation dataset. After that, genetic operators such as crossover and mutation are applied on the parents to generate the offspring population consisting of new models. Then, a selection operation will be performed that selects the better offspring models to be the parents of the next generation, which is often known as survival of the fittest. This selection and reproduction process repeats for a certain number of generations until certain conditions are satisfied. Note that conventional evolutionary NAS needs to evaluate a population of neural networks at each generation and all newly generated neural models are trained from scratch, which are not suited for the online optimization task.

The gradient method became increasingly popular recently, mainly because its search speed is much faster than RL-based and evolutionary NAS methods. In \cite{liu2018darts,dong2019searching} relaxation tricks \cite{trick1992linear} are used to make a weighted sum of candidate operations differentiable so that the gradient descent can be directly employed upon these weights \cite{bottou2010large}. No reinitialization of the model parameters is needed in this approach.

However, the gradient based technique requires much more computation memories than other approaches, since the overall network need to be jointly optimized. To fix this issue, a sampling strategy is proposed by \cite{guo2019single, bender2018understanding}, in which only one or two paths (a path is a sub-network connecting the inputs and outputs) are sampled from a complex neural network for training. In addition, a weight sharing technique \cite{pham2018efficient} is also suggested to avoid reinitialization of the newly sampled models, which saves a plenty of training time. Other techniques such as Bayesian optimization \cite{shahriari2015taking} is also a good approach reducing the computational complexity of NAS \cite{kandasamy2018neural}.

\subsection{Multi-Objective Evolutionary Optimization}
Federated NAS is naturally a multi-objective optimization problem. For instance, the model performance should be maximized, while the payload transferred between the server and clients should be minimized. In the machine learning community, multiple objectives are usually aggregated to be a scalar objective function using hyperparameters. By contrast, the Pareto approach to solving multi-objective optimization problems has been very popular and successful in evolutionary computation \cite{Deb2005}, which has also been extended to machine learning \cite{Jin2006}. The main difference between the Pareto approach and the conventional aggregated approach to machine learning is that in the Pareto approach, no hyperparameters need to be defined and a set of models presenting trade-off relationships between the objectives will be obtained. Finally, one or multiple models can be chosen from the trade-off solutions based on the user's preferences.

The elitist non-dominated sorting genetic algorithm, NSGA-II in short \cite{deb2000fast},  is a very popular multi-objective evolutionary algorithm (MOEA) based on the dominance relationship between the individuals. The overall framework of NSGA-II is summarized in \textbf{Algorithm \ref{nsga2}}.
\begin{algorithm}[htbp]\footnotesize{
\caption{NSGA-II, $N$ is the population size, $t$ is the number of generations}
\algblock{Begin}{End}
\label{nsga2}
\begin{algorithmic}[1]
\State Initialize $t=1$
\State Initialize parents $P(1)$ with a population size $N$
\State Calculate the objective values for $P(1)$
\While {the maximum number of generations is not reached}
\State Generate $N$ offspring $Q(t)$ by applying genetic operators on $P(t)$
\State Calculate objective values of $Q(t)$
\State Combine the parent and offspring populations: $R(t)\leftarrow P(t)+Q(t)$
\State Perform fast non-dominated sorting on $R(t)$ and calculate the crowding distance of all individuals in $R(t)$
\State Sort the combined population according to the dominance relationship and the crowding distance
\State Select the best $N$ individuals from $R(t)$ and store them in parents $P(t+1)$
\State $t\leftarrow t+1$
\EndWhile
\end{algorithmic}}
\end{algorithm}

The main idea of non-dominated sorting is to generate a set of ordered fronts based on the Pareto dominance relationships between the objective values of $R(t)$ and solutions located in the same front cannot dominate with each other. Solutions in the first non-dominated fronts will have a higher priority to be selected. This sorting algorithm has a computation complexity of $O(mN^{2})$, where $m$ is the number of objectives and $N$ is the population size. To promote the solution diversity, a crowding distance that measures the distance between two neighboring solutions in the same front is calculated and those having a large crowding distance have a higher priority to be selected. The computation complexity of crowding distance calculation is $O(mN^{2}logN)$ in the worst case, when all the solutions are located in one non-dominated front. Readers are referred to \cite{deb2000fast} for more details.

\subsection{Online and Offline Evolutionary NAS}
As previously discussed, existing evolutionary NAS methods, also including most RL and gradient based NAS algorithms, are meant for offline model optimization. In other words, they are not suited for scenarios in which neural architecture search must be performed while the network is already in use, such as in recommender systems or vision surveillance systems.

Conventional offline evolutionary NAS is not suited for real-time applications for the following reasons. First, a neural network model is not allowed to be randomly initialized when it is already in use, because random initialization will seriously deteriorate the performance of the neural network. Second, the available computational resource on client devices is limited and therefore, NAS should not require substantially more computational resource. However, EAs are population based search, and at each generation (i.e., time instant), a set of models (depending on the population size) must be assessed, which will considerably increase the computational costs of the clients. Moreover, NAS should not significantly increase the communication costs. This is again challenging for evolutionary NAS, where the parameters of multiple models (depending on the population size) must be transferred between the server and the clients. Finally, the models in offline evolutionary NAS are usually not fully trained to reduce computation time. Thus, they must be trained again at the end of the evolutionary search, which will incur additional communication costs.

\section{Proposed Algorithm}
In this section, we will introduce the encoding method and model structure used in federated NAS at first. And then the two objectives to be optimized and the double-sampling method for evaluating the objectives are explained. Finally, the overall framework of the proposed evolutionary federated NAS is presented.

\subsection{Structure Encoding for Federated NAS}
For real-time NAS, we adopt a light-weighted CNN as the master model, since communication cost is always a primary concern in federated learning. In addition, the search space should not be too large and the total number of layers should be limited to make it appropriate for real-time federated optimization.

\begin{figure}
\centering
\includegraphics[height=7cm, width=8cm]{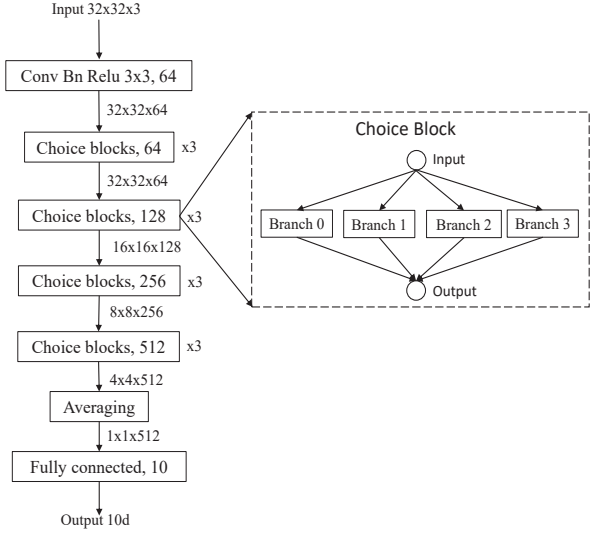}
\caption{An example structure of the master model, in which each choice block consists of four branches.}
\label{dnn}
\end{figure}

\begin{figure}[!t]
\begin{minipage}[t]{1\linewidth}
\centering
\subfigure[Identity Block]{
\begin{minipage}[b]{0.46\textwidth}
\includegraphics[width=1\textwidth]{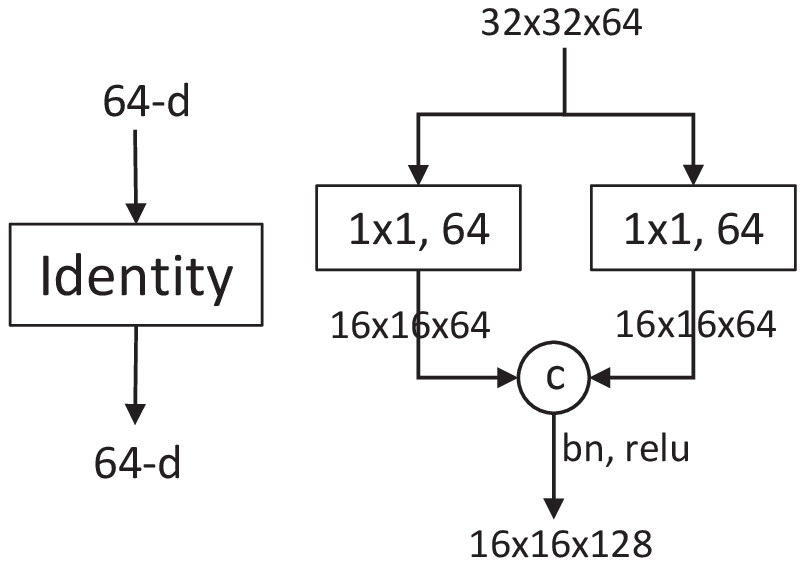}
\end{minipage}
}
\centering
\subfigure[Residual Block]{
\begin{minipage}[b]{0.46\textwidth}
\includegraphics[width=1\textwidth]{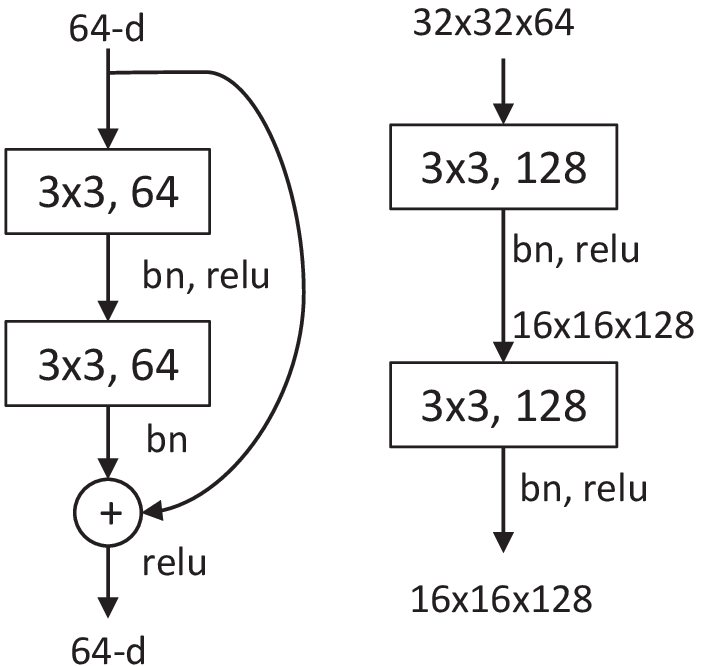}
\end{minipage}
} \\
\centering
\subfigure[Inverted Residual Block]{
\begin{minipage}[b]{0.46\textwidth}
\includegraphics[width=1\textwidth]{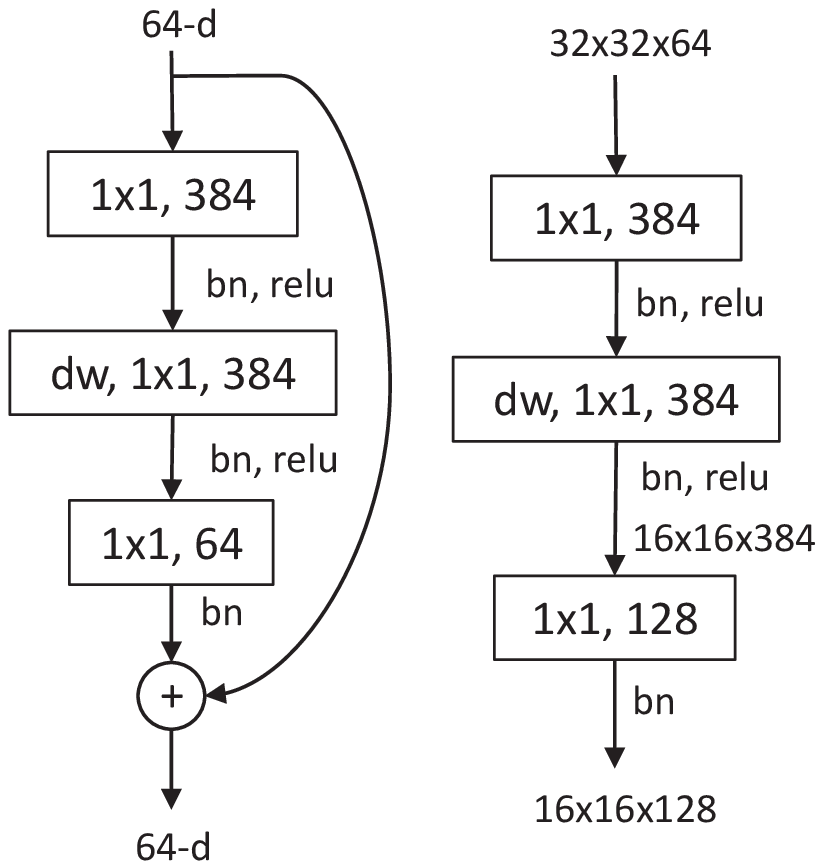}
\end{minipage}
}
\centering
\subfigure[Depth Wise Separate Block]{
\begin{minipage}[b]{0.46\textwidth}
\includegraphics[width=1\textwidth]{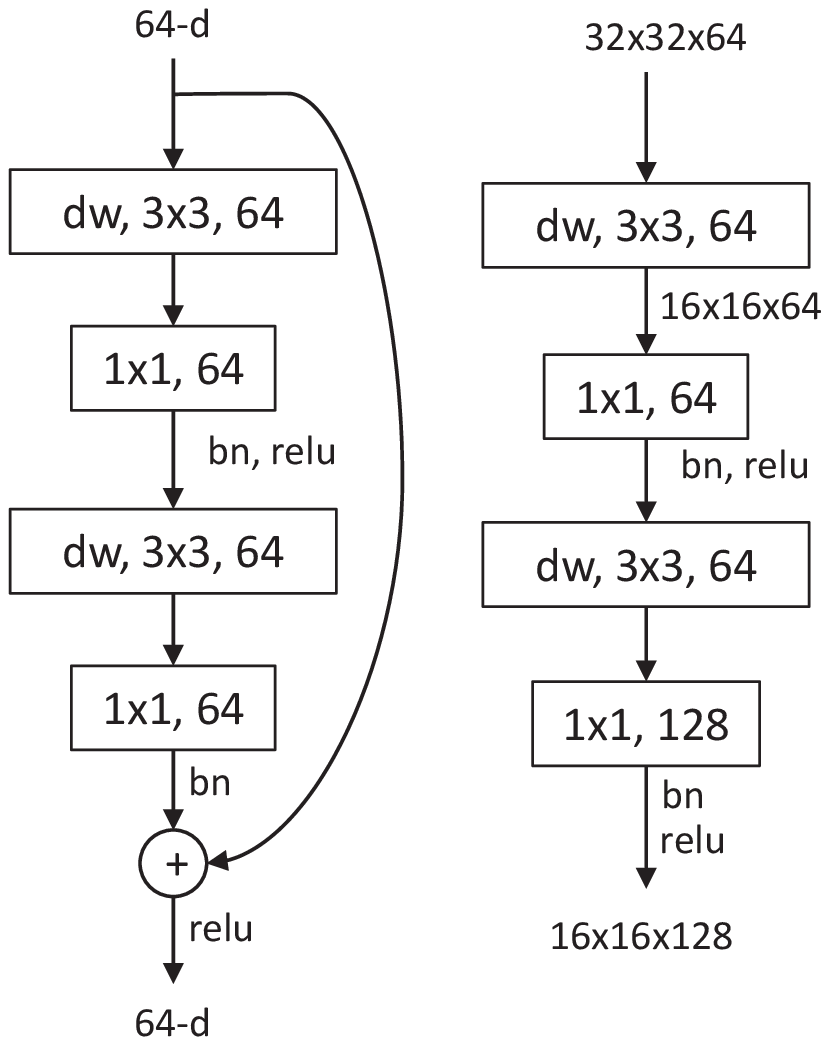}
\end{minipage}
} \\
\caption{Four candidate blocks used in the proposed federated NAS, where the left part of each subfigure is the normal blocks and the right part is the reduction blocks. Symbol $c$ in (a) represents the concatenation operation. Only normal blocks contain shortcut connections.}
\label{cblocks}
\end{minipage}
\end{figure}

The master model used in the proposed online federated NAS is shown in Fig. \ref{dnn}, where a convolutional block, $3\times 4$ choice blocks (each branch of choice block contains two convolutional or more advanced depthwise convolutional layers except for the identity block) and a fully connected layer are linked to build a DNN containing a maximum of $26$ hidden layers. Specifically, the convolutional block consists of three sequentially connected layers with a convolutional layer, a batch normalization layer \cite{ioffe2015batch}, and a rectified linear (ReLu) layer. And one choice block is composed of four pre-defined branches of candidate blocks, namely identity block, residual block, inverted residual block and depthwise separable block, as shown in Fig. \ref{cblocks}. Thus, there are in total $4^{20}$ possible one path sub-networks. In addition, these four candidate blocks are categorized into two groups, one is called normal block whose input and output share the same channel dimension, whereas the other is called reduction block whose output channel dimension is doubled and the spatial dimension is quartered compared to its input.

\emph{Identity block} links its input to its output directly (Fig. \ref{cblocks}(a)), which can be seen as a 'layer removal' operation. For the structure reduction part, it just operates two branches of point-wise convolution with a stride of $2$ at first and then concatenates these two outputs from the channel dimension. As a result, the spatial dimensions of the inputs are quartered and the filter channels are doubled through this identity reduction block.

\emph{Residual block} contains two sequentially connected convolution blocks as shown in Fig. \ref{cblocks}(b), where the normal block has the same structure as the residual block used in ResNet \cite{he2016deep}. Note that the reduction residual block does not contain shortcut connections while the normal block has.

\emph{Inverted residual block} (Fig. \ref{cblocks}(c)) has an 'inverted' bottleneck structure of the residual block, which was first proposed in MobilenetV2 \cite{sandler2018mobilenetv2}. This block contains three convolution layers: 1) a $1\times 1$ expanding point-wise convolution layer, followed by a batch normalization layer and a ReLu activation function. 2) a $3\times 3$ depthwise convolution layer \cite{chollet2017xception, howard2017mobilenets}, followed by a batch normalization layer and a ReLu activation function. 3) a $1\times 1$ spatially filtered point-wise convolution layer followed by a batch normalization layer.

The intuition of using expanding at the first layer instead of in the last layer as done in the bottleneck layer is that a nonlinear activation function such as ReLu may cause layer information loss \cite{han2017deep} and using a nonlinear projection upon a high-dimensional space can mitigate this issue. After the tensor is mapped back to a low-dimensional space again through the last point-wise layer, the ReLu function is not needed to prevent information loss.

\emph{Depthwise separable block} consists of two depthwise convolution operations \cite{chollet2017xception} (Fig. \ref{cblocks}(d)) that incurs lower computation cost than the conventional convolution operation. It has been proved in \cite{howard2017mobilenets} that $3\times 3$ depthwise convolution consumes about one-eighth to one-ninth of the computation time of the standard convolution at the expense of a small deterioration in performance.

\begin{figure}
\centering
\includegraphics[height=5cm, width=8cm]{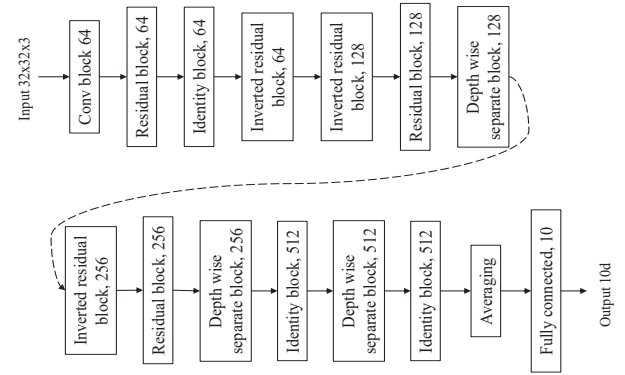}
\caption{A sub-model represented by the choice key $[0, 1, 0, 0, 1, 0, 1, 0, 0, 1, 1, 1, 1, 0, 0, 1, 1, 1, 0, 0, 1, 1, 0, 0]$.}
\label{smodel}
\end{figure}

For each communication round, only one branch of all $12$ choice blocks is sampled from the master model and then downloaded to a client device for reducing communication costs and computation resources required at the local device. This sampled sub-model can be encoded into a two-bit binary string with a total length of $2\times 12=24$ bits. Every two bits in the code represent one specific branch in the choice block. For instance, $[0, 0]$ represents branch $0$, which is the identity block, $[0, 1]$ represents branch $1$, i.e., the residual block, $[1, 0]$ represents branch $2$, the inverted residual block, and $[1, 1]$ represents branch $3$, which is the depthwise separable block. Therefore, the binary string (also called the choice key) $[0, 1, 0, 0, 1, 0, 1, 0, 0, 1, 1, 1, 1, 0, 0, 1, 1, 1, 0, 0, 1, 1, 0, 0]$ can be decoded into a sub-model with structure as shown in Fig. \ref{smodel}.

\subsection{A Double-Sampling for Objective Evaluations}
Offline evolutionary optimization is intrinsically not suited for federated learning. Although one or multiple light weighted models with high performance can be found by an offline evolutionary algorithm in the last generation \cite{zhu2019multi}, a large amount of extra computation and communication resources is required. For instance, for an EA of a population size of $N$, each client in an offline evolutionary NAS algorithm must evaluate the fitness of $N$ individuals at each generation, which is $N-1$ times in extra compared to the gradient method. In addition, the models are repeatedly randomly reinitialized and trained from scratch on the clients, which may not be good enough for use during the optimization.

In order to address the above issues, a double-sampling technique is proposed here to develop a real-time evolutionary federated NAS, in which at each generation, the global model for each individual are sub-sampled from a common \textit{master model}, whereas the clients for training the global model of one individual is sub-sampled from the participating clients. Specifically, a choice key is generated for each individual to randomly sample a sub-network from the master model to be this individual's global model. Then, the global model of this individual (a sampled sub-model) will be downloaded to a randomly sampled subset of the participating clients. The number of clients to be chosen, say $L$, for training one global model of an individual is determined by the ratio between the number of individuals and the number of participating clients, i.e., $L = \lfloor m/N \rfloor$, where $m=CK$ is the number of participating clients, $K$ is the total number of clients and $C$ is the participation ratio at the current round. Here, we assume that the number of clients is equal to or larger than that of the population size.

In this work, two objectives are to be optimized: one is the test error of each global model and the other is the floating point operations per second (FLOPs) of the model. Recall that the global model of each individual is a sub-model of the master model sampled using a choice key. Population initialization of the proposed methods is composed of the following four main steps.
\begin{itemize}
\item Initialize the mater model. Generate the initial parent population containing $N$ individuals, each representing a sub-model sampled from the master model using a choice key. Sample $L$ clients for each individual without replacement. That is, each client should be sampled only once.
\item Download the sub-model of each parent individual to the $L$ selected clients and train it using the data on the clients. Once the training is completed, upload the $L$ local sub-models to the server for aggregation to update the master model.
\item Generate $N$ offspring individuals using crossover and mutation. Similarly, generate a choice key for each offspring individual to sample a sub-model from the master model. Download the sub-model of each offspring individual to $L$ randomly sampled participating clients and train it on these clients. Note that the weights of the sub-models are inherited directly from the master model and will not be reinitialized in training. Upload the trained local sub-models and aggregate them to update the master model.
\item Finally, download the master model together with the choice keys of all parents and offspring to all clients to evaluate the test errors and FLOPs. Upload the test errors and FLOPs to the server and calculate the weighted averaging of test errors for each individual.
\end{itemize}

Once the objective values of all parent and offspring individuals are calculated, environmental selection can be carried out to generate the parent individuals of the next generation based on the elitist non-dominated sorting and crowding distance, as discussed in Algorithm 2 in Section II.C.

In the following generations, similar steps as described above will be carried out, except that the master model is updated only once at each generation after the global model (also a sub-model randomly sampled from the master model) of all offspring individuals is trained on the sampled clients and the resulting local sub-models are aggregated. It should be emphasized that at each generation, the master model shared by all individuals need to be downloaded to all participating clients only once for fitness evaluations.

\begin{figure}
\centering
\includegraphics[height=5.5cm, width=7.5cm]{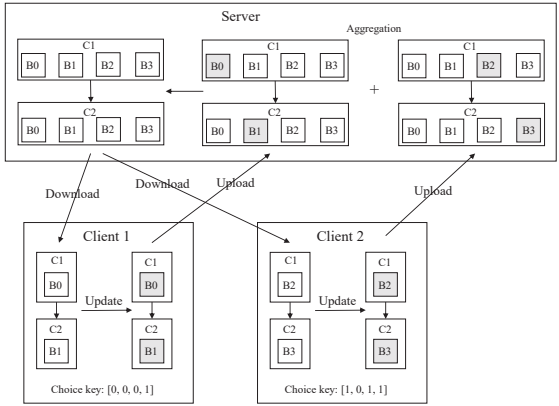}
\caption{An illustrative example of model aggregation. The master model contains two choice blocks ($C_1$, $C_2$) and $B_0$, $B_1$, $B_2$, $B_3$ are four different branches in a choice block. The two sampled sub-models are downloaded to client 1 and client 2, respectively for training. After the updated sub-models are uploaded, they are are filled with the remaining sub-models (those not updated in this round for this individual) to reconstruct the master model before all reconstructed master models are aggregated.}
\label{fnas}
\end{figure}

Note that model aggregation is different from that in the conventional federated learning. The reason is that different clients may be sampled for different individuals, and different individuals may have different model structures, which cannot be directly aggregated. Fig. \ref{fnas} shows an illustrative example, where the master model has two choice blocks $C_1$ and $C_2$. There are two individuals, and the choice keys for the two individuals are [0, 0, 0, 1] and [1, 0, 1, 1], respectively. The resulting sub-models are $B_0$ and $B_1$, and $B_2$ and $B_3$. We further assume that client 1 is chosen for training sub-model $B_0$ and $B_1$ and client 2 for $B2$ and $B3$. Then each client updates its model parameters according to the available local dataset and then upload the trained model to the server, which is denoted by the shaded square. And then two master models are reconstructed by filling in the sub-models that are not updated, which are denoted by the white squares. Since the reconstructed master models have the same structure, they can be easily aggregated using i.e., the weighted averaging. The pseudo code for model aggregation is presented in Algorithm \ref{serveragg}.

The advantage of the above filling and aggregation method is able to prevent abrupt changes in some sub-models and improve convergence in federated learning. In addition, this aggregation method does not require extra communication resources, since this operation is only performed on the server.

From the above description, we can see that the double-sampling strategy fits perfectly well with the population based evolutionary search so that the objective values of all individuals at one generation can be evaluated within one communication round, seamlessly embedding a generation of architecture search into one round of training in federated learning.

\begin{algorithm}[htbp]\footnotesize{
\caption{Model Aggregation, $m$ is the total number of client uploads. $k$ is the client index. $I$ is the total number of hidden layers of the master model, $i$ is the hidden layer index of the learning model, $B$ is the total branches in one choice block, $b$ is the branch block index, $n$ is the number of data samples on all clients, $n_{k}$ is the number of data on client $k$, and $t$ is the communication round.}
\algblock{Begin}{End}
\label{serveragg}
\begin{algorithmic}[1]
\State $\theta(t-1)$ is the parameters of the master model in the last communication round, $\theta_{b}^{i}(t)$ is the parameters of the master model for the $b$-th branch in the $i$-th hidden layer.
\State Receive client model parameters $\theta_{k}$ and choice key $C_{k}$, where $\theta_{k}^{i}$ is the model parameters and $C_{k}^{i}$ is the choice of $i$-th hidden layer.
\State \textbf{Server Aggregation:}
\State Let $\theta(t)\leftarrow 0$
\For{each $i\in I$}
\For{each $k\in m$}
\If{$\theta_{k}^{i}$ is not in choice blocks}
\State $\theta^{i}(t)\leftarrow \sum_{k=1}^{m}\frac{n_{k}}{n}\theta _{k}^{i}$
\Else
\For{each branch $b\in B$}
\If $C_{k}^{i}==b$
\State $\theta_{b}^{i}(t)\leftarrow \theta_{b}^{i}(t)+\frac{n_{k}}{n}\theta_{k}^{i}$
\Else
\State $\theta_{b}^{i}(t)\leftarrow \theta_{b}^{i}(t)+\frac{n_{k}}{n}\theta_{b}^{i}(t-1)$
\EndIf
\EndFor
\EndIf
\EndFor
\EndFor
\State \textbf{Return} $\theta(t)$
\end{algorithmic}}
\end{algorithm}

\subsection{Overall Framework}
\begin{figure*}
\centering
\includegraphics[scale=0.5]{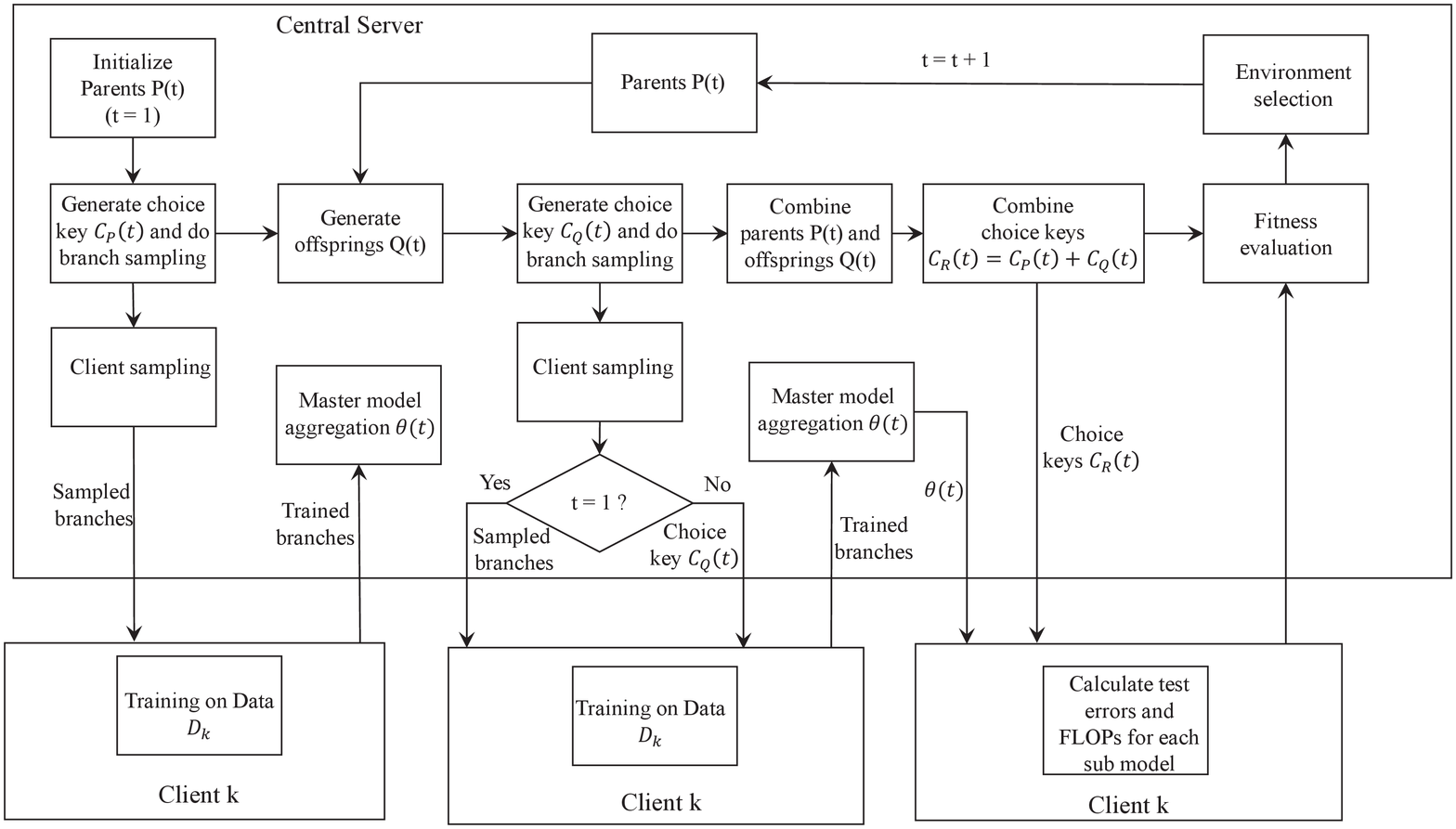}
\caption{The overall framework for multi-objective online evolutionary federated NAS, where $t$ is the communication round, $P(t)$ is the parent population and $C_P(t)$ is the corresponding choice keys, $Q(t)$ is the offspring population and $C_Q(t)$ is its choice keys, $R(t)$ is the combined population and $C_R(t)$ is the choice keys of the combined population, $\theta(t)$ is all trainable parameters of the master model.}
\label{moeafnasframe}
\end{figure*}

We use NSGA-II to optimize both model FLOPs and performances for the online evolutionary federated NAS framework. The framework is illustrated in Fig.~\ref{moeafnasframe} and the pseudo code is listed in Algorithm \ref{moeafnas}.

\begin{algorithm}[htbp]\footnotesize{
\caption{Online Federated NAS by NSGA-II, $N$ is the population size, $t$ is the number of generations, $K$ indicates the total numbers of clients; $B$ is size of mini-batch, $ E\ $ is equal to training iterations and $ \eta \ $ is the learning rate, $n$ is the total number of data points on all clients, $n_{k}$ is the number of data points on client $k$}
\algblock{Begin}{End}
\label{moeafnas}
\begin{algorithmic}[1]
\State // \textbf{\emph{Double sampling method used here}}
\State \textbf{Server: }
\State Initialize $ {\theta(0)}\ $
\State $\theta(t) \leftarrow  \theta(0)$
\State $t \leftarrow 1$
\State // \textbf{\emph{Server master model sampling (model sampling)}}
\State Randomly sub sample parent choice keys $C_P(t)$ with a population size $N$
\For {each communication round $ t = 1,2,...\ $}
\State // \textbf{\emph{For online optimization, the generation is equal to the communication round}}
\State Convert all $C_P(t)$ choice keys into binary codes $Cb_P(t)$
\State Generate $Cb_Q(t)$ with the size of $N$ by binary genetic operators
\State Convert $Cb_Q(t)$ into offspring choice keys $C_Q(t)$
\State $C_R(t) \leftarrow C_P(t)+C_Q(t)$
\State Select $ m = C \times K $ clients, $ C \in (0,1) $ clients
\If{$t\leq 1$}
\State Generate sub models $\theta^{p} \in \theta(t)$, $p \in C_P(t)$
\State // \textbf{\emph{Client sampling (clients sampling)}}
\State Randomly sub sample $m$ clients to $N$ groups
\For {$p \in C_P(t)$, $g \in N$ ($|C_P(t)|=N$)}
\State Download $\theta^{P}$ to all clients in group $g$
\EndFor
\For {each client $ k \in m\ $}
\State Wait \textbf{Client $k$ Update} for synchronization
\State $\theta(t) \leftarrow$ \emph{Do server aggregation in} \textbf{Algorithm \ref{serveragg}}
\EndFor
\EndIf
\State // \textbf{\emph{No need to reinitialize the model parameters of offspring models}}
\State Generate sub models $\theta^{q} \in \theta(t)$, $q \in C_Q(t)$
\State // \textbf{\emph{Client sampling (clients sampling)}}
\State Randomly sub sample $m$ clients to $N$ groups
\For {$q \in C_Q(t)$, $g \in N$ ($|C_Q(t)|=N$)}
\If{$t > 1$}
\State Download the choice key $q$ to all clients in group $g$
\Else
\State Download $\theta^{q}$ and the choice key $q$ to all clients in group $g$
\EndIf
\EndFor
\For {each client $ k \in m\ $}
\State Wait \textbf{Client $k$ Update} for synchronization
\State $\theta(t) \leftarrow$ \emph{Do server aggregation in} \textbf{Algorithm \ref{serveragg}}
\EndFor
\State // \textbf{\emph{Do NSGA-II optimization}}
\State Calculate FLOPs of all sub models in $C_R(t)$
\For {each client $ k \in m\ $}
\State Download master model $\theta(t)$ and choice keys $C_R(t)$ to client $k$
\State Calculate the test errors for all sub models in $C_R(t)$
\State Upload them to the server
\EndFor
\State Do weighted averaging on test errors of all uploads based on the local data size to achieve the final test errors of all sub models in $C_R(t)$
\State Do fast non dominated sorting
\State Do crowding distance sorting
\State // \textbf{\emph{New solutions are generated within each communication round}}
\State Generate new parent choice keys $C_P(t)$ from $C_R(t)$
\State $t \leftarrow t+1$
\EndFor \\
\State \textbf{Client $k$ Update: }
\If{Receive one choice key $q$}
\State Sub sample ${\theta(t)}$ based on the choice key to generate ${\theta ^k}$
\Else
\State $ {\theta ^k} \leftarrow \theta^{p} or \theta^{q}$
\EndIf
\For {each iteration from 1 to $E$}
\For {batch $ b \in B\ $}
\State $ {\theta ^k} = {\theta ^k} - \eta \nabla {L_k}({\theta ^k},b)\ $
\EndFor
\EndFor
\State return ${\theta ^k}$ to the server
\end{algorithmic}}
\end{algorithm}

The fitness evaluations are done for both parent and offspring populations at every generation, which is equivalent to a communication round in the proposed real-time evolutionary NAS. For fitness evaluations, both the master model and all choice keys $C_R(t)$ are downloaded to all participating clients. Thus, we do not need to download any model parameters for training of any sub-models in the next round and it is sufficient to download the choice keys only (refer to Lines 32-33 in  Algorithm \ref{moeafnas}. After that, each client uploads the updated local sub-models to the server for model aggregation. As a result, the proposed model sampling method can reduce both local computation complexity and communication costs for uploading the models.

It should be noticed that the parent sub-models are trained only at the first generation. In the subsequent evolutionary optimization, only offspring sub-models need to be trained. However, all $2N$ sub-models sampled by $C_R(t)$ need to be evaluated to calculate the test errors for fitness evaluations at each generation, because training the offspring sub-models will also affect the parameters of the parent sub-model since parent and offspring sub-models always share weights of the master model. In addition, we do not need to reinitialize the model parameters of the sampled offspring sub-models before training starts. Due to the client sampling strategy, the population size $N$ does not affect the communication costs for fitness evaluations, since the entire master model is downloaded from the server and sampling of the master model can be done on the clients.

Since NSGA-II is a Pareto based multi-objective optimization algorith which can find a set of optimal models that trade off between accuracy and computational complexity (FLOPs). Therefore, for online applications, the user needs to articulate preferences to select one of the Pareto optial solutions from the parent population in each round. In practice, the Pareto solutions with the high test accuracies or those near the knee points \cite{8638825,8477885,6975108} are preferred unless there are other strong user-specified preferences.

\section{Experimental Results}
\subsection{NSGA-II Settings}
The settings for NSGA-II are listed in the Table \ref{nsga2p}. Here, we use binary one-point crossover and binary bit flip mutation as the genetic operators. Since the master network only contains four choice blocks, a two-bit binary string is used for representation.

\begin{table}[]
\centering
\caption{Parameters of NSGA-II}
\begin{tabular}{ll}
\hline
Parameter             & Value         \\
\hline
Generations           & 500           \\
Population            & 10        \\
Crossover Probability & 0.9           \\
Mutation Probability  & 0.1           \\
Bit Length & 2 \\
\hline
\end{tabular}
\label{nsga2p}
\end{table}

\subsection{Federated Learning Settings}
The hyper parameters for federated learning are presented in Table \ref{flp}.

\begin{table}[]
\centering
\caption{Hyper parameters for federated learning}
\begin{tabular}{ll}
\hline
Parameter & value \\
\hline
Total Clients         & 10, 20, 50 \\
Local Epochs          & 1          \\
Train Batch Size      & 50         \\
Test Batch Size       & 100        \\
Initial Learning rate & 0.1        \\
Momentum              & 0.5        \\
Learning Decay        & 0.995   \\
\hline
\end{tabular}
\label{flp}
\end{table}

Here, learning decay means the decay of the learning rate over each communication round. Apart from this, the number of total communication rounds is not set here, because it is equal to the number of generations in real-time evolutionary optimization.

\subsection{Model and Dataset Settings}
The master model used in this work is a deep neural network with multiple branches containing a total of 28 layers (12 choice blocks, each containing two convolution layers). The number of channels for all block layers are [64, 64, 64, 128, 128, 128, 256, 256, 256, 512, 512, 512]. The overall structure of the master model is shown in Fig. \ref{dnn}. Apart from this, the trainable parameters in the batch normalization layer may slow down the convergence speed in federated learning, since they perform poorly for learning with small batch sizes \cite{ioffe2017batch} and weight sharing training paradigms \cite{yu2018slimmable,guo2019single,chu2019fairnas}, especially for \emph{non-IID} scenarios. In addition, we disable both the variables and exponential moving average variables in the batch normalization layer because we found that they may cause the divergence of the master model.

We adopt Cifar10 \cite{Krizhevsky} as the dataset, which contains 50000 training and 10000 testing 32x32 RGB images with 10 different kinds of objects. For \emph{IID} federated simulations, all training image data are evenly and randomly distributed to each local client without overlaps. For experiments on \emph{non-IID} data, each client has images with $5$ different kinds of objects. We do not consider very extreme cases where each client has data with only $1$ or $2$ classes, since this is not realistic in the real world environment. For instance, it is not beneficial to collaboratively train a global model with clients having completely different datasets.

Note that we do not apply any data augmentation \cite{perez2017effectiveness} in our simulation, since the server cannot do any operations on the client data in federated learning.

In all experiments, we use one GTX 1080Ti GPU.

\subsection{Experiment Results}

\subsubsection{Baseline Model Used in Federated Learning}
We use ResNet18 as the baseline model and its parameters are provided in Table \ref{resnet}. As previously mentioned, all trainable parameters in the batch normalization layers are removed, resulting in a total FLOPs (MAC) of 0.5587G on the cifar10 dataset. The settings for the federated learning follow those given in Section III.
\begin{table}[]
\centering
\caption{Architecture of ResNet for cifar10}
\begin{tabular}{lll}
\toprule
Layer Name   & Output Size & 18 Layers \\
\toprule
Conv1        & 32x32       & 3x3, 64       \\
\midrule
Conv2\_x     & 32x32       & $\begin{bmatrix}3\times 3, 64\\3\times 3, 64\end{bmatrix}\times 2$          \\
\midrule
Conv3\_x     & 16x16       & $\begin{bmatrix}3\times 3, 128\\3\times 3, 128\end{bmatrix}\times 2$          \\
\midrule
Conv4\_x     & 8x8         & $\begin{bmatrix}3\times 3, 256\\3\times 3, 256\end{bmatrix}\times 2$          \\
\midrule
Conv5\_x     & 4x4         & $\begin{bmatrix}3\times 3, 512\\3\times 3, 512\end{bmatrix}\times 2$          \\
\midrule
Average Pool & 1x1               \\
\bottomrule
\end{tabular}
\label{resnet}
\end{table}

\subsection{Federated Evolutionary NAS Results}
\begin{figure}[!t]
\begin{minipage}[t]{1\linewidth}
\centering
\subfigure[10 clients, \emph{IID} data]{
\begin{minipage}[b]{0.46\textwidth}
\includegraphics[width=1\textwidth]{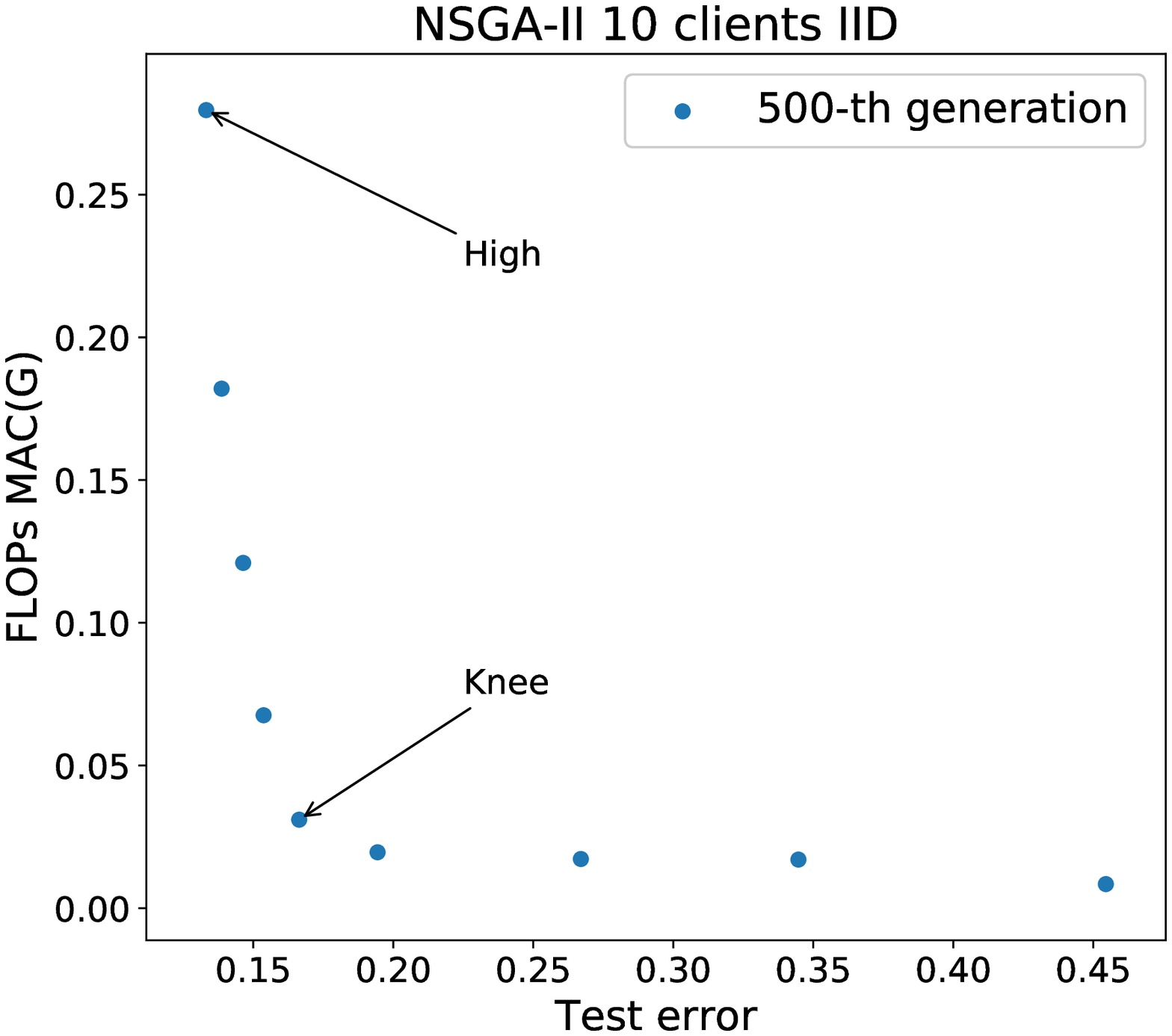}
\end{minipage}
}
\centering
\subfigure[10 clients, \emph{non-IID} data]{
\begin{minipage}[b]{0.46\textwidth}
\includegraphics[width=1\textwidth]{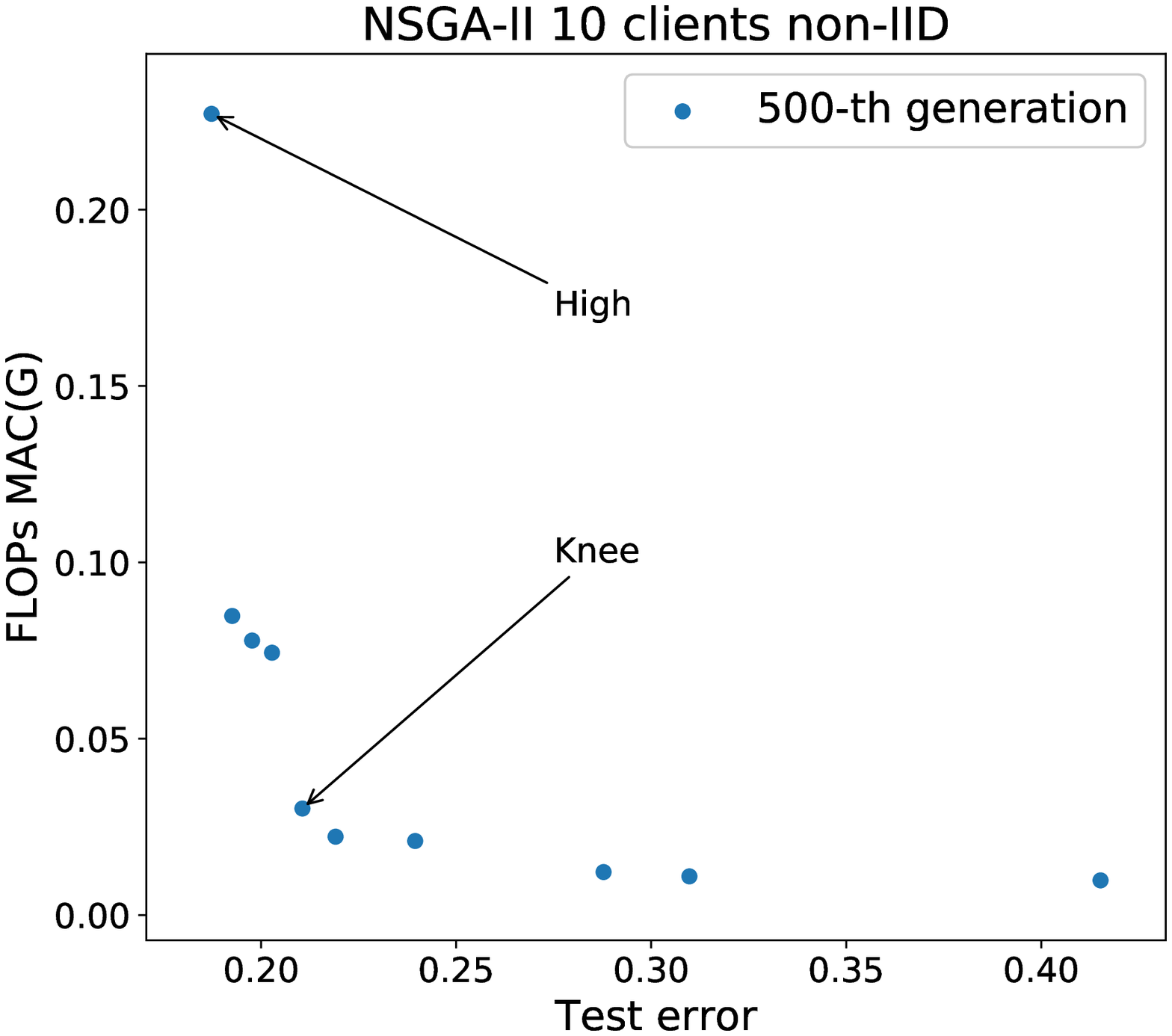}
\end{minipage}
} \\
\centering
\subfigure[20 clients, \emph{IID} data]{
\begin{minipage}[b]{0.46\textwidth}
\includegraphics[width=1\textwidth]{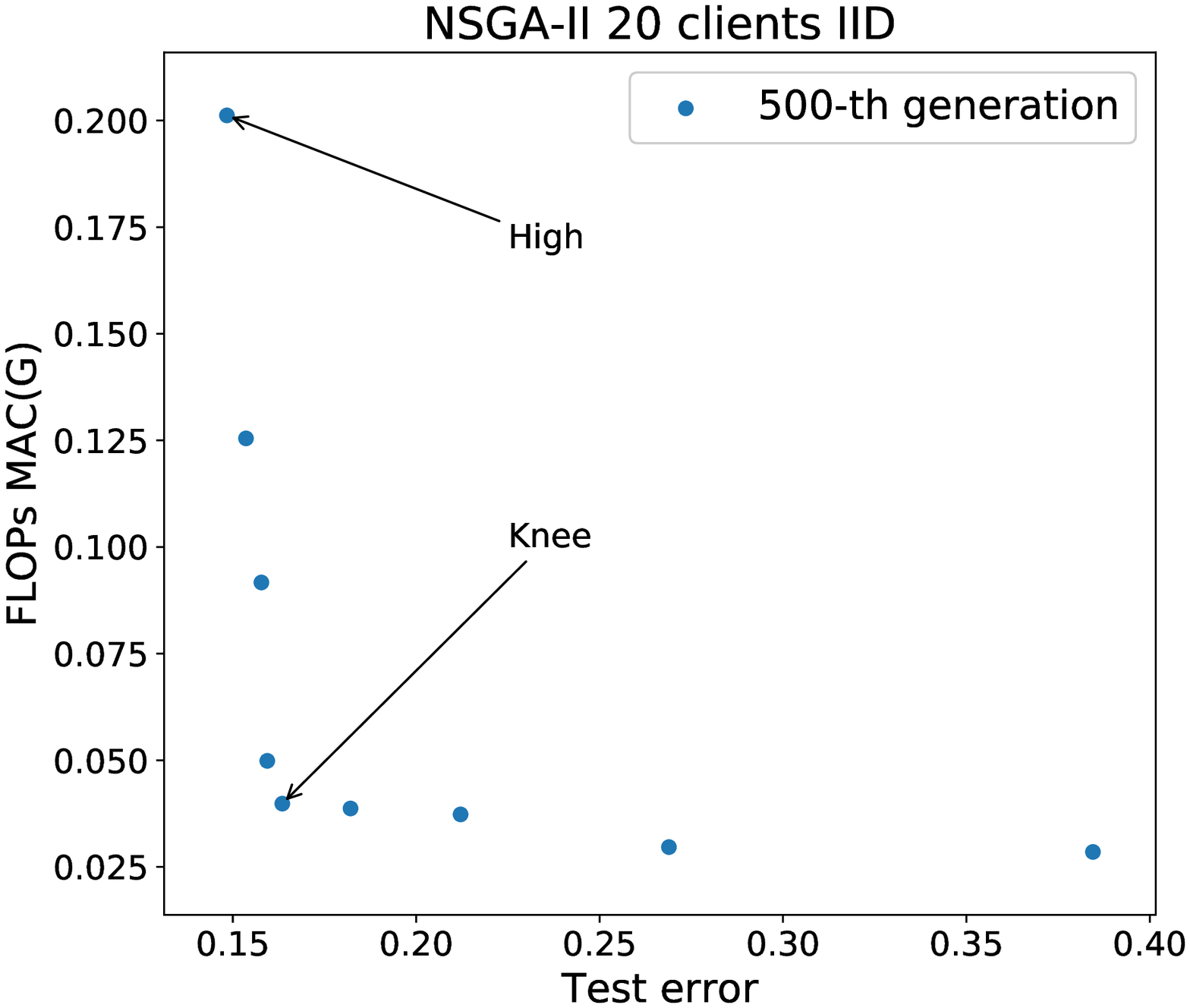}
\end{minipage}
}
\centering
\subfigure[20 clients, \emph{non-IID} data]{
\begin{minipage}[b]{0.46\textwidth}
\includegraphics[width=1\textwidth]{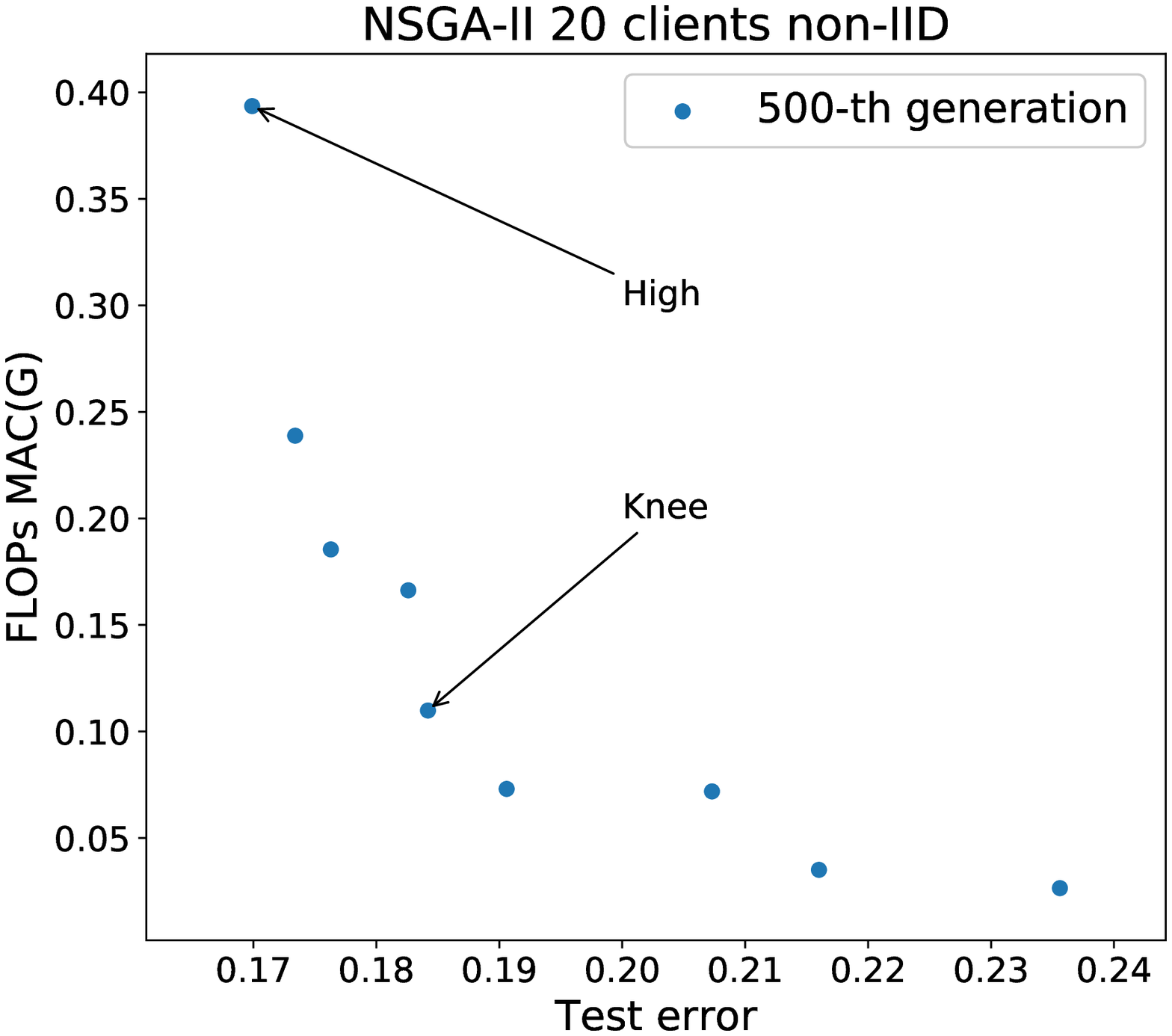}
\end{minipage}
} \\
\centering
\subfigure[50 clients, \emph{IID} data]{
\begin{minipage}[b]{0.46\textwidth}
\includegraphics[width=1\textwidth]{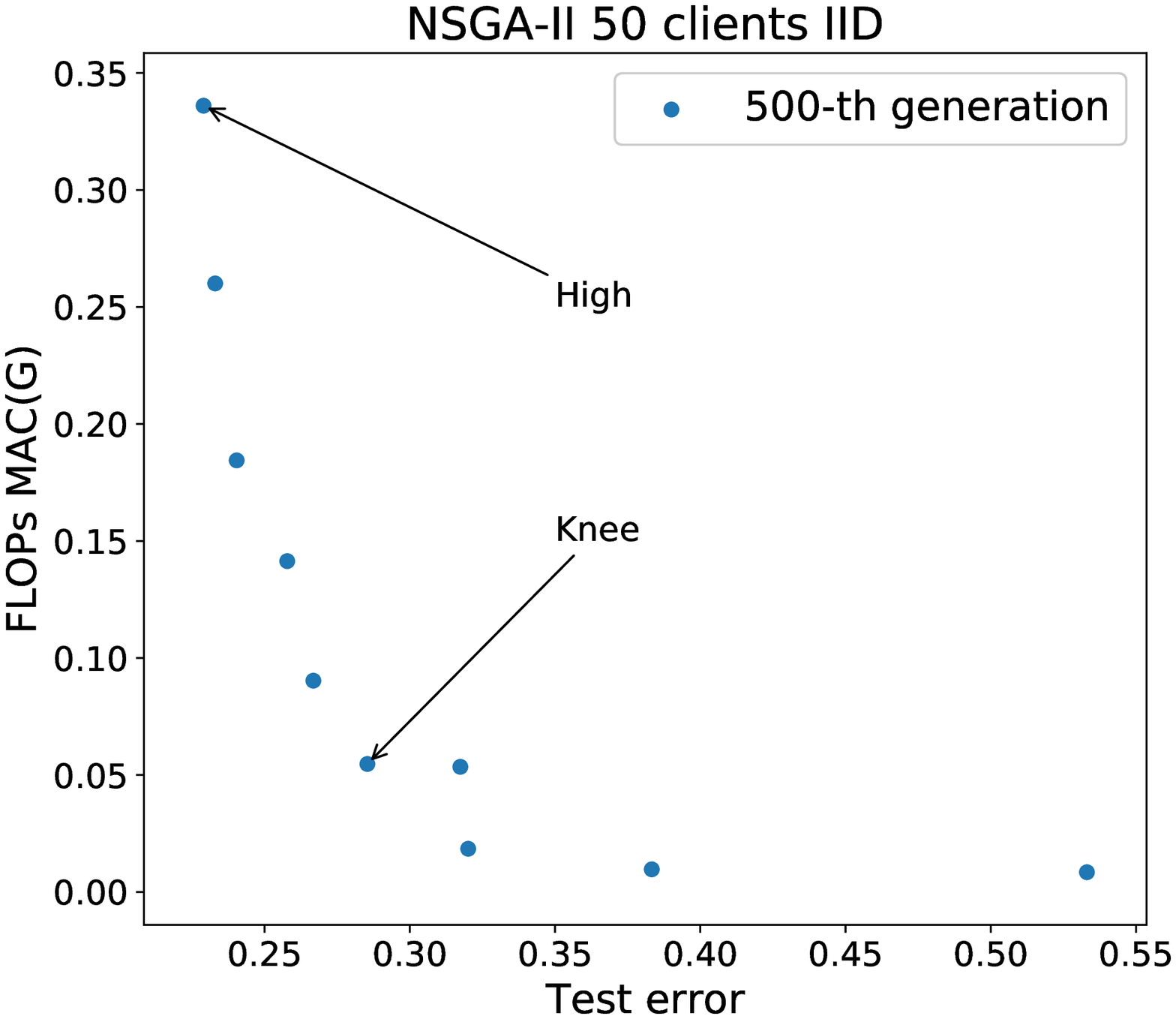}
\end{minipage}
}
\centering
\subfigure[50 clients, \emph{non-IID} data]{
\begin{minipage}[b]{0.46\textwidth}
\includegraphics[width=1\textwidth]{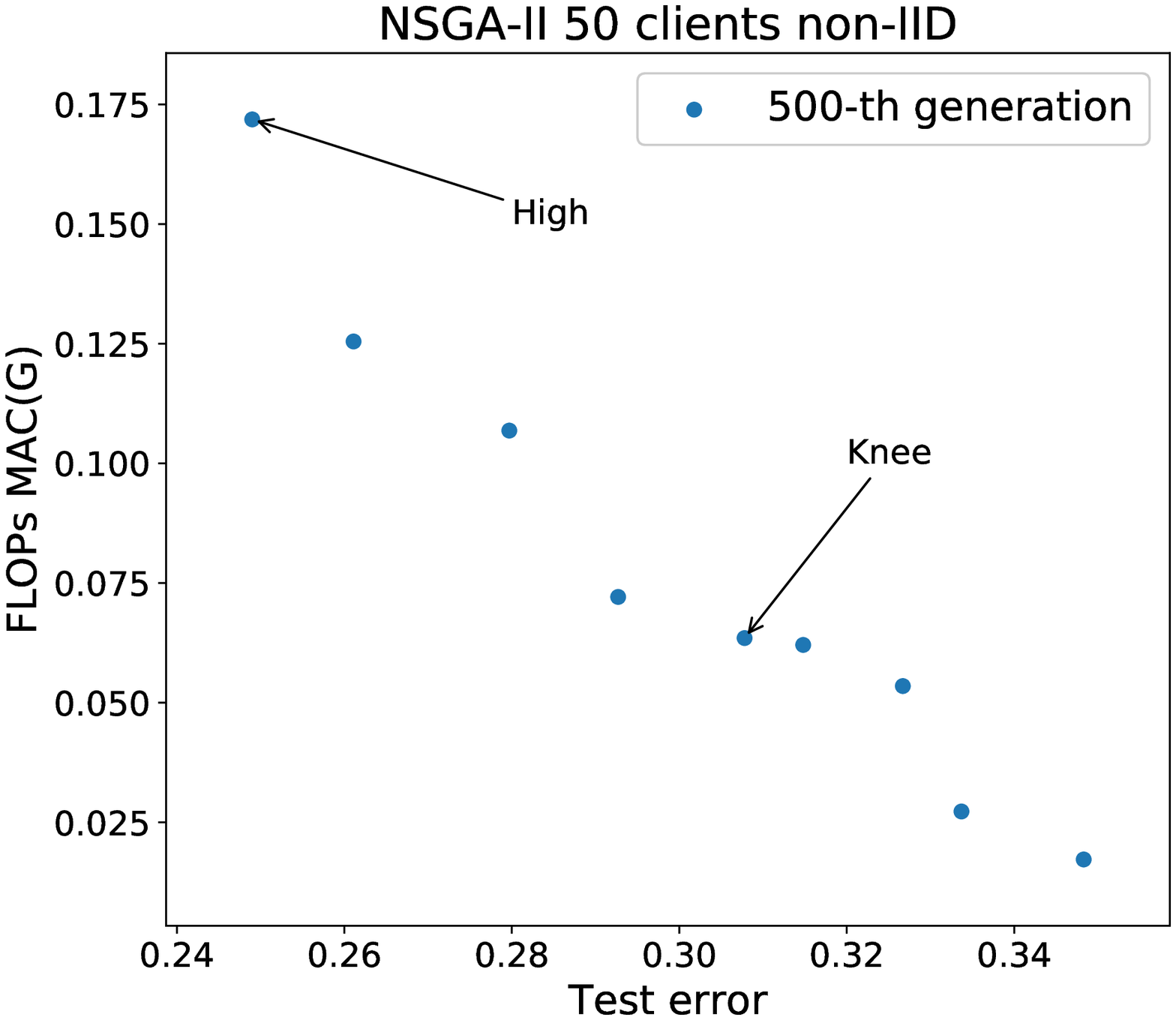}
\end{minipage}
} \\
\caption{Pareto optimal solutions obtained on the \emph{IID} and \emph{non-IID} data for $10$, $20$ and $50$ clients.}
\label{nsgac}
\end{minipage}
\end{figure}

We adopt three different numbers of clients (10, 20 and 50) for real-time multi-objective evolutionary federated NAS and the obtained Pareto optimal solutions on both \emph{IID} and \emph{non-IID} data after 500 rounds (generations) of optimization are shown in Fig. \ref{nsgac}. From these results we can see that the real-time evolutionary federated NAS algorithm is able to achieve a set of evenly distributed Pareto optimal solutions.

The following two observations can be made. First, the classification accuracies of the optimized models on the \emph{IID} data are better than those for on the \emph{non-IID} data. Second, the smaller the number of the clients, the better the classification performance. These two phenomenons are reasonable, since learning on \emph{IID} data is much easier than learning on \emph{non-IID} data in federated learning. On the other hand, the more the number of clients, the less data there is on each client, as the amount of data in total is given.

To take a closer look at the obtained models, we present the test accuracies and FLOPs of the model having the highest test accuracy (called \textbf{High}) and the knee solution (called \textbf{Knee}), as indicated in Fig. \ref{nsgac}, as well as that of the ResNet in Table \ref{nsga2t}.

\begin{table}[]
\centering
\caption{Comparison between ResNet and NSGA-II evolved two pareto solutions}
\begin{tabular}{ccccc}
\toprule
Model            & clients                  & IID & Test Accuracy & FLOPs (MAC) \\
\toprule
ResNet            & 10                 & Yes & 81.14\%       & 0.5587G     \\
High  & 10   & Yes & \textbf{86.68\%}       & 0.2796G     \\
Knee   & 10   & Yes & 83.36\%       & \textbf{0.031G }     \\
ResNet           & 10                  & No  & 80.29\%       & 0.5587G     \\
High & 10  & No  & \textbf{81.27\%}       & 0.2272G     \\
Knee  & 10    & No  & 78.94\%       & \textbf{0.0302G} \\
\midrule
ResNet              & 20               & Yes & 82.5\%       & 0.5587G     \\
High  & 20 & Yes & \textbf{85.16\%}       & 0.2012G     \\
Knee  & 20    & Yes & 83.65\%       & \textbf{0.0398G }     \\
ResNet                & 20             & No  & 79.91\%       & 0.5587G     \\
High  & 20 & No  & \textbf{83.01\%}       & 0.3936G     \\
Knee   & 20   & No  & 81.58\%       & \textbf{0.1098G} \\
\midrule
ResNet       & 50                      & Yes & \textbf{80.78\%}       & 0.5587G     \\
High  & 50 & Yes & 77.1\%       & 0.3360G     \\
Knee   & 50    & Yes & 71.46\%       & \textbf{0.0547G }     \\
ResNet                  & 50           & No  & \textbf{77.63\% }      & 0.5587G     \\
High  & 50  & No  & 75.1\%       & 0.1719G     \\
Knee    & 50    & No  & 69.22\%       & \textbf{0.0635G} \\
\bottomrule
\end{tabular}
\label{nsga2t}
\end{table}

From these results, we can see that federated learning with a total of $10$ and $20$ clients, the model having the highest accuracy found by the proposed algorithm is better than the original ResNet in accuracy and has a much lower computational complexity. For 10 clients, the knee solutions found by the proposed algorithm have a much lower model FLOPS than the original ResNet. By contrast for 20 clients, the model FLOPs of the knee solutions is again must lower than that of the ResNet and the performance on both \emph{IID} and \emph{non-IID} data is better. However, for $50$ clients, the test accuracies of the models found by the proposed algorithm is worse than the ResNet, although the model FLOPs are lower. This indicates that it becomes harder to find an optimal global model as the amount of data on each client becomes less.

\subsection{Real-time Performance}
\begin{figure}[!t]
\begin{minipage}[t]{1\linewidth}
\centering
\subfigure[$20$ clients, \emph{IID} data]{
\begin{minipage}[b]{0.46\textwidth}
\includegraphics[width=1\textwidth]{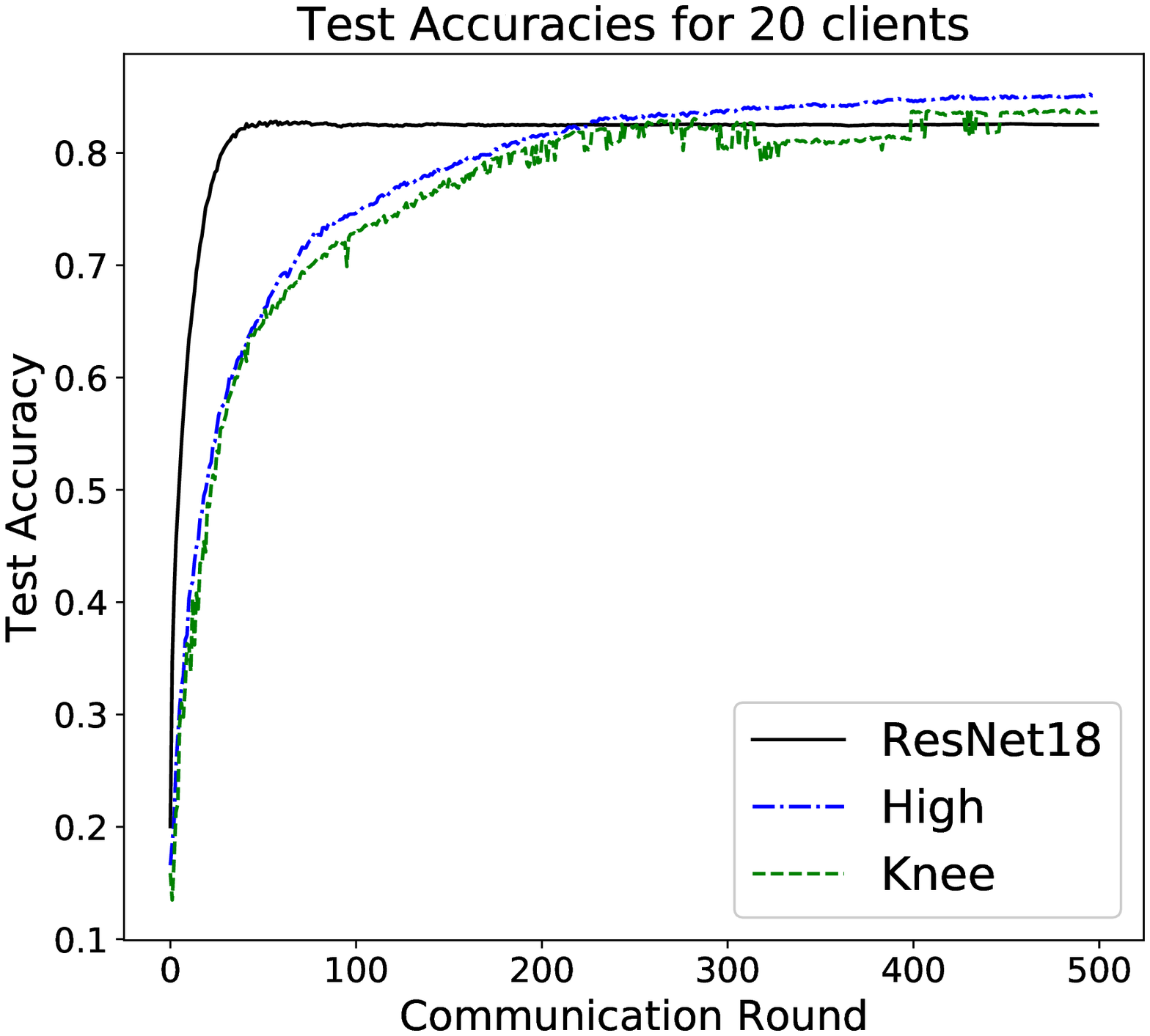}
\end{minipage}
}
\centering
\subfigure[$20$ clients, \emph{non-IID} data]{
\begin{minipage}[b]{0.46\textwidth}
\includegraphics[width=1\textwidth]{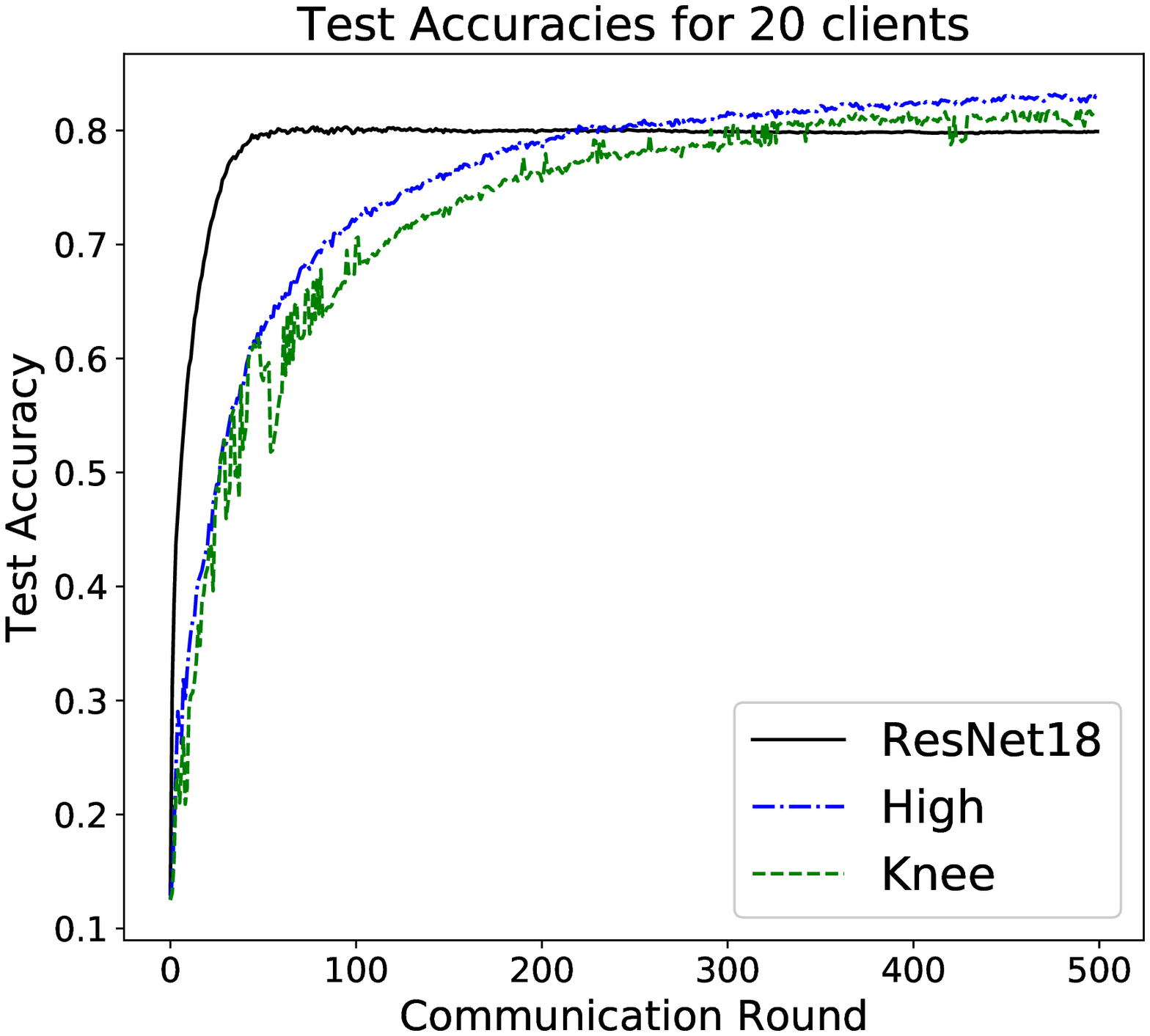}
\end{minipage}
} \\
\centering
\subfigure[$20$ clients, \emph{IID} data]{
\begin{minipage}[b]{0.46\textwidth}
\includegraphics[width=1\textwidth]{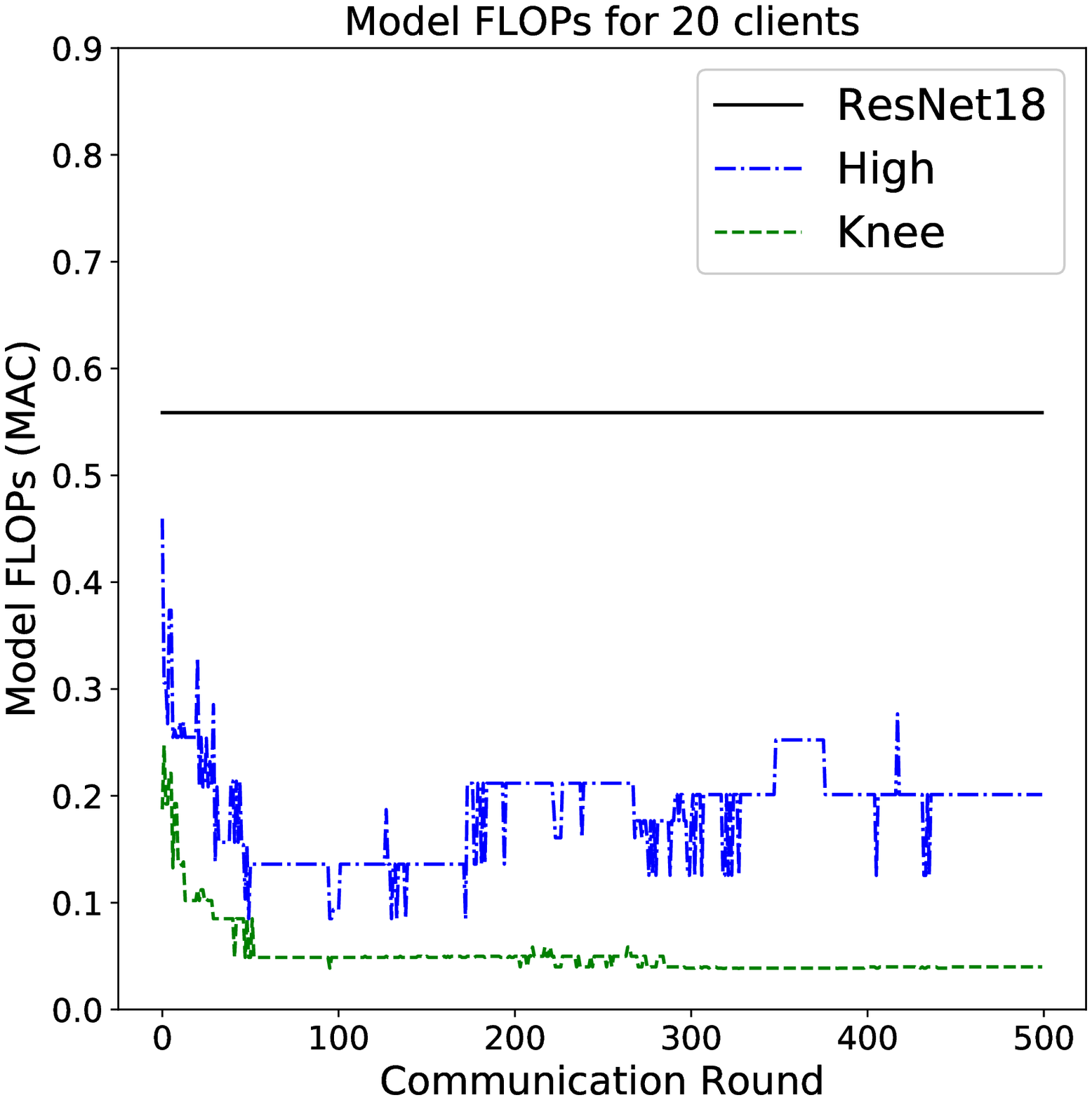}
\end{minipage}
}
\centering
\subfigure[$20$ clients, \emph{non-IID} data]{
\begin{minipage}[b]{0.46\textwidth}
\includegraphics[width=1\textwidth]{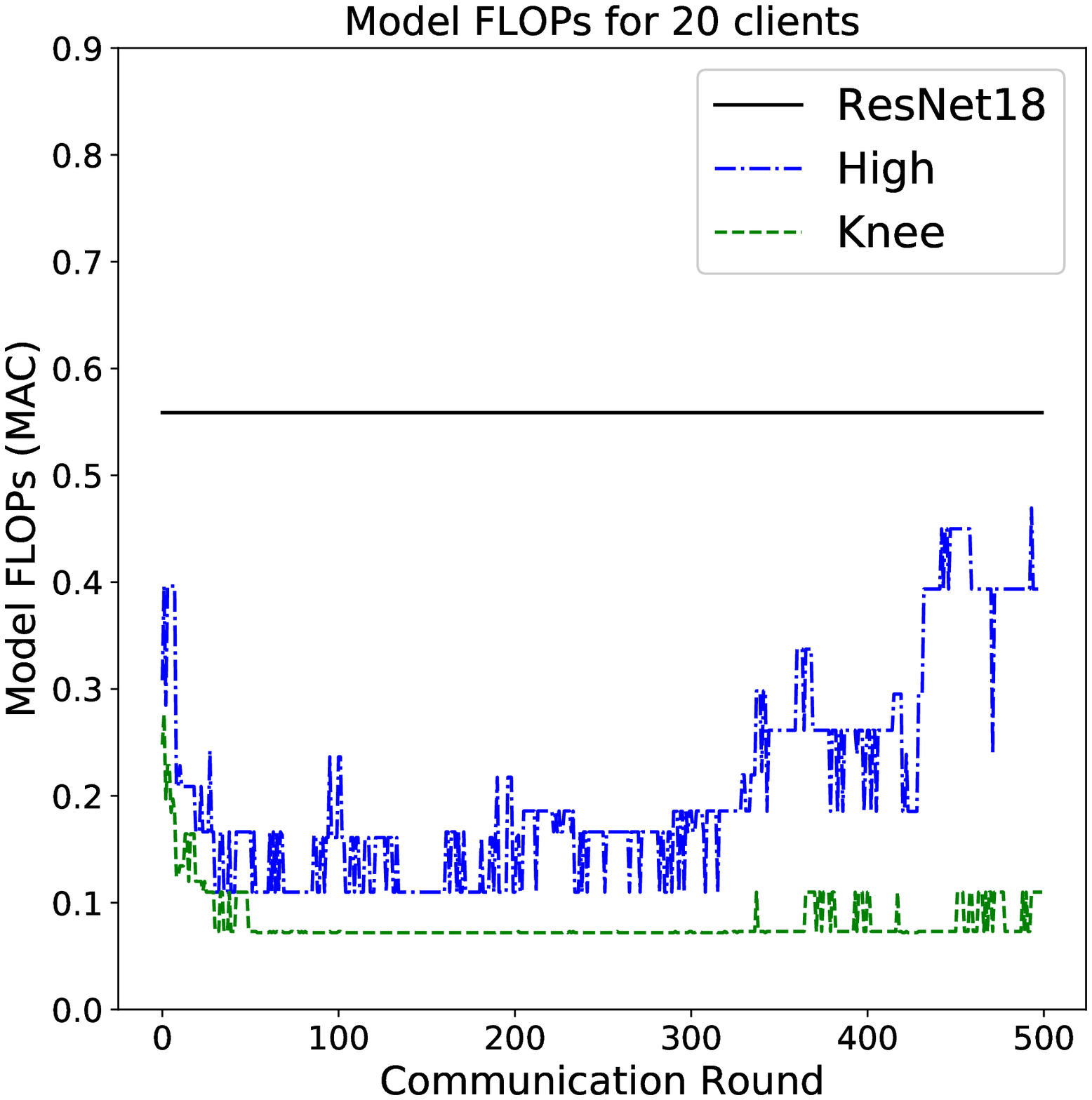}
\end{minipage}
} \\
\caption{Test accuracies and the model FLOPs of the best model and the knee solution in each round (generation) of the evolutionary search.}
\label{ci20}
\end{minipage}
\end{figure}

Since the proposed method is meant for real-time purposes, here we examine the performance of two models during the optimization, one is the model having the highest test accuracy and the other is the knee solution. For simplicity, we only investigate the real-time performance when the number of participating clients is $20$, which is presented in Fig. \ref{ci20}. For comparison, the performance of ResNet18 is also plotted. We can see clearly that ResNet18 performs better than both models found by the evolutionary search method at the early stage. However, the two solutions are able to outperform ResNet after approximately $200$ communication rounds. We can also find that the performance of the best model is very stable during the evolutionary search, although the knee solution experiences some minor fluctuations in performance. Both models perform much more stably than those in the conventional offline evolutionary NAS in \cite{zhu2019multi}.

The model FLOPs of the two solutions are shown in Fig. \ref{ci20}(c)(d), which are smaller than that of the original ResNet18.

From the above results, we can see that the proposed real-time evolutionary NAS algorithm is not only able to find light weighted models, but also ensure stable and competitive performance during the optimization.

\subsection{Comparison with Offline Federated NAS}
Here, we also compare the proposed algorithm with the offline evolutionary federated NAS optimization. In the conventional offline model, the sampled model is downloaded to all clients with the same structure and offspring models should be reinitialized and trained from scratch, similar to the settings in \cite{zhu2019multi}.

\begin{figure}
\centering
\includegraphics[height=5cm, width=6cm]{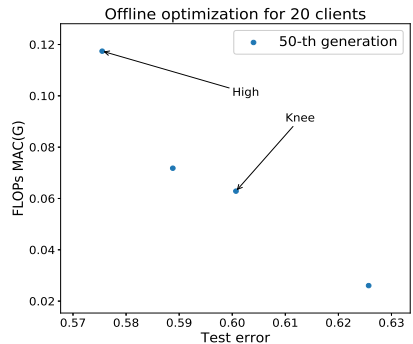}
\caption{Pareto optimal solutions found by the offline evolutionary federated NAS.}
\label{offpf}
\end{figure}

\begin{figure}
\centering
\includegraphics[height=5cm, width=6cm]{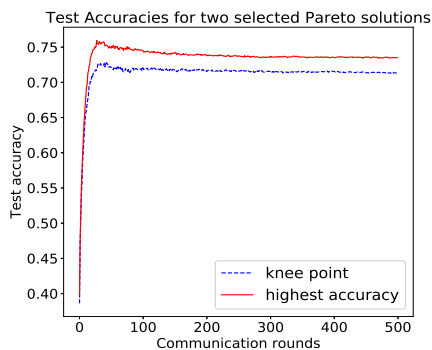}
\caption{Test accuracies of the two selected solutions found by the offline evolutionary NAS.}
\label{offacc}
\end{figure}

For a fair comparison, models are collaboratively trained with one communication round for fitness evaluations. The number of clients is set to be $20$ and the Pareto optimal solutions obtained on the \emph{non-IID} data after 50 generations are plotted in Fig. \ref{offpf}. Only four solutions are found in the last generation. Finally, the model having the highest accuracy and the knee solution as indicated in Fig. \ref{offpf} are selected to be trained from scratch for $500$ communication rounds and their learning curve is shown in Fig. \ref{offacc}. From these results we can see that the performance of the best model found by the offline evolutionary NAS is lower than 75$\%$ in this case, which is attributed to the very fast convergence in the early stage of the evolutionary search. These results also imply that in the offline evolutionary NAS, it is non-trivial to define the number of training rounds at each generation.

It takes about $16.1$ hrs/gpu for the real-time evolutionary NAS to run 500 generations (rounds). In contrast, the offline evolutionary NAS for 50 generations needs about $6.55$ hrs/gpu without considering the re-training time. This means that on average, the real-time evolutionary NAS is approximately five times faster than the offline evolutionary NAS algorithm for each round of search.

\section{Conclusions and Future Work}
This paper proposes a real-time multi-objective evolutionary method for federated NAS, which can effectively avoid extra communication costs and computational resources and maintain a stable performance of the models during the optimization. This is achieved by a double-sampling approach that samples a master model shared by all individuals in the same population and samples the participating clients for training the global model. This way, one generation of evolutionary optimization can be embedded and completed within one single communication round, thereby reducing both communication and computation costs.

The experimental results demonstrate that the proposed evolutionary federated NAS framework is able to find a set of evenly distributed Pareto optimal solutions for both \emph{IID} and \emph{non-IID} datasets. Among these Pareto optimal solutions, we can obtain models having different architectures that present a trade-off between classification performance and computational complexity. In addition, we show that these models are computationally much simpler than the standard model while the performance is still highly competitive. The high computational efficiency of the proposed real-time evolutionary NAS is also confirmed by the comparative results with the conventional offline evolutionary method.

The present work is a first valuable step towards the application of neural architecture search to the federated learning framework. In the future, we are going to verify and extend the the proposed algorithm for real-time NAS in large-scale federated learning systems. In addition, new techniques remain to be developed to deal with data that are vertically partitioned and distributed on the clients.


%

%
%
%
%
%

\ifCLASSOPTIONcaptionsoff
  \newpage
\fi


%



{\footnotesize\bibliography{ref}

\providecommand{\noopsort}[1]{}\providecommand{\singleletter}[1]{#1}%
\begin{thebibliography}{10}

\bibitem{mcmahan2016communication}
H.~B. McMahan, E.~Moore, D.~Ramage, S.~Hampson, {\em et~al.},
  ``Communication-efficient learning of deep networks from decentralized
  data,'' {\em arXiv preprint arXiv:1602.05629}, 2016.

\bibitem{shokri2015privacy}
R.~Shokri and V.~Shmatikov, ``Privacy-preserving deep learning,'' in {\em
  Proceedings of the 22nd ACM SIGSAC conference on computer and communications
  security}, pp.~1310--1321, ACM, 2015.

\bibitem{konevcny2016federated}
J.~Kone{\v{c}}n{\`y}, H.~B. McMahan, F.~X. Yu, P.~Richt{\'a}rik, A.~T. Suresh,
  and D.~Bacon, ``Federated learning: Strategies for improving communication
  efficiency,'' {\em arXiv preprint arXiv:1610.05492}, 2016.

\bibitem{caldas2018expanding}
S.~Caldas, J.~Kone{\v{c}}ny, H.~B. McMahan, and A.~Talwalkar, ``Expanding the
  reach of federated learning by reducing client resource requirements,'' {\em
  arXiv preprint arXiv:1812.07210}, 2018.

\bibitem{han2015deep}
S.~Han, H.~Mao, and W.~J. Dally, ``Deep compression: Compressing deep neural
  networks with pruning, trained quantization and huffman coding,'' {\em arXiv
  preprint arXiv:1510.00149}, 2015.

\bibitem{chen2019communication}
Y.~Chen, X.~Sun, and Y.~Jin, ``Communication-efficient federated deep learning
  with asynchronous model update and temporally weighted aggregation,'' {\em
  arXiv preprint arXiv:1903.07424}, 2019.

\bibitem{zhu2019multi}
H.~Zhu and Y.~Jin, ``Multi-objective evolutionary federated learning,'' {\em
  IEEE transactions on neural networks and learning systems}, 2019.

\bibitem{zoph2016neural}
B.~Zoph and Q.~V. Le, ``Neural architecture search with reinforcement
  learning,'' {\em arXiv preprint arXiv:1611.01578}, 2016.

\bibitem{pham2018efficient}
H.~Pham, M.~Y. Guan, B.~Zoph, Q.~V. Le, and J.~Dean, ``Efficient neural
  architecture search via parameter sharing,'' {\em arXiv preprint
  arXiv:1802.03268}, 2018.

\bibitem{suganuma2017genetic}
M.~Suganuma, S.~Shirakawa, and T.~Nagao, ``A genetic programming approach to
  designing convolutional neural network architectures,'' in {\em Proceedings
  of the Genetic and Evolutionary Computation Conference}, pp.~497--504, ACM,
  2017.

\bibitem{real2019regularized}
E.~Real, A.~Aggarwal, Y.~Huang, and Q.~V. Le, ``Regularized evolution for image
  classifier architecture search,'' in {\em Proceedings of the AAAI Conference
  on Artificial Intelligence}, vol.~33, pp.~4780--4789, 2019.

\bibitem{real2017large}
E.~Real, S.~Moore, A.~Selle, S.~Saxena, Y.~L. Suematsu, J.~Tan, Q.~V. Le, and
  A.~Kurakin, ``Large-scale evolution of image classifiers,'' in {\em
  Proceedings of the 34th International Conference on Machine Learning-Volume
  70}, pp.~2902--2911, JMLR. org, 2017.

\bibitem{liu2017hierarchical}
H.~Liu, K.~Simonyan, O.~Vinyals, C.~Fernando, and K.~Kavukcuoglu,
  ``Hierarchical representations for efficient architecture search,'' {\em
  arXiv preprint arXiv:1711.00436}, 2017.

\bibitem{liu2018darts}
H.~Liu, K.~Simonyan, and Y.~Yang, ``Darts: Differentiable architecture
  search,'' {\em arXiv preprint arXiv:1806.09055}, 2018.

\bibitem{dong2019searching}
X.~Dong and Y.~Yang, ``Searching for a robust neural architecture in four gpu
  hours,'' in {\em Proceedings of the IEEE Conference on Computer Vision and
  Pattern Recognition}, pp.~1761--1770, 2019.

\bibitem{Sun2019}
Y.~Sun, H.~Wang, B.~Xue, Y.~Jin, G.~G. Yen, and M.~Zhang, ``Surrogate-assisted
  evolutionary deep learning using an end-to-end random forest-based
  performance predictor,'' {\em IEEE Transactions on Evolutionary Computation},
  2019.
\newblock DOI: 10.1109/TEVC.2019.2924461.

\bibitem{tan2019mnasnet}
M.~Tan, B.~Chen, R.~Pang, V.~Vasudevan, M.~Sandler, A.~Howard, and Q.~V. Le,
  ``Mnasnet: Platform-aware neural architecture search for mobile,'' in {\em
  Proceedings of the IEEE Conference on Computer Vision and Pattern
  Recognition}, pp.~2820--2828, 2019.

\bibitem{pan2009survey}
S.~J. Pan and Q.~Yang, ``A survey on transfer learning,'' {\em IEEE
  Transactions on knowledge and data engineering}, vol.~22, no.~10,
  pp.~1345--1359, 2009.

\bibitem{torrey2010transfer}
L.~Torrey and J.~Shavlik, ``Transfer learning,'' in {\em Handbook of research
  on machine learning applications and trends: algorithms, methods, and
  techniques}, pp.~242--264, IGI Global, 2010.

\bibitem{ioffe2015batch}
S.~Ioffe and C.~Szegedy, ``Batch normalization: Accelerating deep network
  training by reducing internal covariate shift,'' {\em arXiv preprint
  arXiv:1502.03167}, 2015.

\bibitem{ammad2019federated}
M.~Ammad-ud din, E.~Ivannikova, S.~A. Khan, W.~Oyomno, Q.~Fu, K.~E. Tan, and
  A.~Flanagan, ``Federated collaborative filtering for privacy-preserving
  personalized recommendation system,'' {\em arXiv preprint arXiv:1901.09888},
  2019.

\bibitem{he2016deep}
K.~He, X.~Zhang, S.~Ren, and J.~Sun, ``Deep residual learning for image
  recognition,'' in {\em Proceedings of the IEEE conference on computer vision
  and pattern recognition}, pp.~770--778, 2016.

\bibitem{bottou1991stochastic}
L.~Bottou, ``Stochastic gradient learning in neural networks,'' {\em
  Proceedings of Neuro-N{\i}mes}, vol.~91, no.~8, p.~12, 1991.

\bibitem{lecun2015deep}
Y.~LeCun, Y.~Bengio, and G.~Hinton, ``Deep learning,'' {\em nature}, vol.~521,
  no.~7553, pp.~436--444, 2015.

\bibitem{lecun1995convolutional}
Y.~LeCun, Y.~Bengio, {\em et~al.}, ``Convolutional networks for images, speech,
  and time series,'' {\em The handbook of brain theory and neural networks},
  vol.~3361, no.~10, p.~1995, 1995.

\bibitem{krizhevsky2012imagenet}
A.~Krizhevsky, I.~Sutskever, and G.~E. Hinton, ``Imagenet classification with
  deep convolutional neural networks,'' in {\em Advances in neural information
  processing systems}, pp.~1097--1105, 2012.

\bibitem{szegedy2016rethinking}
C.~Szegedy, V.~Vanhoucke, S.~Ioffe, J.~Shlens, and Z.~Wojna, ``Rethinking the
  inception architecture for computer vision,'' in {\em Proceedings of the IEEE
  conference on computer vision and pattern recognition}, pp.~2818--2826, 2016.

\bibitem{huang2017densely}
G.~Huang, Z.~Liu, L.~Van Der~Maaten, and K.~Q. Weinberger, ``Densely connected
  convolutional networks,'' in {\em Proceedings of the IEEE conference on
  computer vision and pattern recognition}, pp.~4700--4708, 2017.

\bibitem{szegedy2017inception}
C.~Szegedy, S.~Ioffe, V.~Vanhoucke, and A.~A. Alemi, ``Inception-v4,
  inception-resnet and the impact of residual connections on learning,'' in
  {\em Thirty-First AAAI Conference on Artificial Intelligence}, 2017.

\bibitem{szegedy2015going}
C.~Szegedy, W.~Liu, Y.~Jia, P.~Sermanet, S.~Reed, D.~Anguelov, D.~Erhan,
  V.~Vanhoucke, and A.~Rabinovich, ``Going deeper with convolutions,'' in {\em
  Proceedings of the IEEE conference on computer vision and pattern
  recognition}, pp.~1--9, 2015.

\bibitem{zoph2018learning}
B.~Zoph, V.~Vasudevan, J.~Shlens, and Q.~V. Le, ``Learning transferable
  architectures for scalable image recognition,'' in {\em Proceedings of the
  IEEE conference on computer vision and pattern recognition}, pp.~8697--8710,
  2018.

\bibitem{baker2016designing}
B.~Baker, O.~Gupta, N.~Naik, and R.~Raskar, ``Designing neural network
  architectures using reinforcement learning,'' {\em arXiv preprint
  arXiv:1611.02167}, 2016.

\bibitem{zhong2018practical}
Z.~Zhong, J.~Yan, W.~Wu, J.~Shao, and C.-L. Liu, ``Practical block-wise neural
  network architecture generation,'' in {\em Proceedings of the IEEE Conference
  on Computer Vision and Pattern Recognition}, pp.~2423--2432, 2018.

\bibitem{schuster1997bidirectional}
M.~Schuster and K.~K. Paliwal, ``Bidirectional recurrent neural networks,''
  {\em IEEE Transactions on Signal Processing}, vol.~45, no.~11,
  pp.~2673--2681, 1997.

\bibitem{angeline1994evolutionary}
P.~J. Angeline, G.~M. Saunders, and J.~B. Pollack, ``An evolutionary algorithm
  that constructs recurrent neural networks,'' {\em IEEE transactions on Neural
  Networks}, vol.~5, no.~1, pp.~54--65, 1994.

\bibitem{yao1999evolving}
X.~Yao, ``Evolving artificial neural networks,'' {\em Proceedings of the IEEE},
  vol.~87, no.~9, pp.~1423--1447, 1999.

\bibitem{trick1992linear}
M.~A. Trick, ``A linear relaxation heuristic for the generalized assignment
  problem,'' {\em Naval Research Logistics (NRL)}, vol.~39, no.~2,
  pp.~137--151, 1992.

\bibitem{bottou2010large}
L.~Bottou, ``Large-scale machine learning with stochastic gradient descent,''
  in {\em Proceedings of COMPSTAT'2010}, pp.~177--186, Springer, 2010.

\bibitem{guo2019single}
Z.~Guo, X.~Zhang, H.~Mu, W.~Heng, Z.~Liu, Y.~Wei, and J.~Sun, ``Single path
  one-shot neural architecture search with uniform sampling,'' {\em arXiv
  preprint arXiv:1904.00420}, 2019.

\bibitem{bender2018understanding}
G.~Bender, ``Understanding and simplifying one-shot architecture search,''
  2018.

\bibitem{shahriari2015taking}
B.~Shahriari, K.~Swersky, Z.~Wang, R.~P. Adams, and N.~De~Freitas, ``Taking the
  human out of the loop: A review of bayesian optimization,'' {\em Proceedings
  of the IEEE}, vol.~104, no.~1, pp.~148--175, 2015.

\bibitem{kandasamy2018neural}
K.~Kandasamy, W.~Neiswanger, J.~Schneider, B.~Poczos, and E.~P. Xing, ``Neural
  architecture search with bayesian optimisation and optimal transport,'' in
  {\em Advances in Neural Information Processing Systems}, pp.~2016--2025,
  2018.

\bibitem{Deb2005}
K.~Deb, {\em Multi-objective Optimization using Evolutionary Algorithms}.
\newblock Wiley, 2005.

\bibitem{Jin2006}
Y.~Jin, ed., {\em Multi-objective Machine Learning}.
\newblock Springer, 2006.

\bibitem{deb2000fast}
K.~Deb, S.~Agrawal, A.~Pratap, and T.~Meyarivan, ``A fast elitist non-dominated
  sorting genetic algorithm for multi-objective optimization: Nsga-ii,'' in
  {\em International conference on parallel problem solving from nature},
  pp.~849--858, Springer, 2000.

\bibitem{sandler2018mobilenetv2}
M.~Sandler, A.~Howard, M.~Zhu, A.~Zhmoginov, and L.-C. Chen, ``Mobilenetv2:
  Inverted residuals and linear bottlenecks,'' in {\em Proceedings of the IEEE
  Conference on Computer Vision and Pattern Recognition}, pp.~4510--4520, 2018.

\bibitem{chollet2017xception}
F.~Chollet, ``Xception: Deep learning with depthwise separable convolutions,''
  in {\em Proceedings of the IEEE conference on computer vision and pattern
  recognition}, pp.~1251--1258, 2017.

\bibitem{howard2017mobilenets}
A.~G. Howard, M.~Zhu, B.~Chen, D.~Kalenichenko, W.~Wang, T.~Weyand,
  M.~Andreetto, and H.~Adam, ``Mobilenets: Efficient convolutional neural
  networks for mobile vision applications,'' {\em arXiv preprint
  arXiv:1704.04861}, 2017.

\bibitem{han2017deep}
D.~Han, J.~Kim, and J.~Kim, ``Deep pyramidal residual networks,'' in {\em
  Proceedings of the IEEE Conference on Computer Vision and Pattern
  Recognition}, pp.~5927--5935, 2017.

\bibitem{8638825}
G.~{Yu}, Y.~{Jin}, and M.~{Olhofer}, ``Benchmark problems and performance
  indicators for search of knee points in multiobjective optimization,'' {\em
  IEEE Transactions on Cybernetics}, pp.~1--14, 2019.

\bibitem{8477885}
G.~{Yu}, Y.~{Jin}, and M.~{Olhofer}, ``A method for a posteriori identification
  of knee points based on solution density,'' in {\em 2018 IEEE Congress on
  Evolutionary Computation (CEC)}, pp.~1--8, July 2018.

\bibitem{6975108}
X.~{Zhang}, Y.~{Tian}, and Y.~{Jin}, ``A knee point-driven evolutionary
  algorithm for many-objective optimization,'' {\em IEEE Transactions on
  Evolutionary Computation}, vol.~19, pp.~761--776, Dec 2015.

\bibitem{ioffe2017batch}
S.~Ioffe, ``Batch renormalization: Towards reducing minibatch dependence in
  batch-normalized models,'' in {\em Advances in neural information processing
  systems}, pp.~1945--1953, 2017.

\bibitem{yu2018slimmable}
J.~Yu, L.~Yang, N.~Xu, J.~Yang, and T.~Huang, ``Slimmable neural networks,''
  {\em arXiv preprint arXiv:1812.08928}, 2018.

\bibitem{chu2019fairnas}
X.~Chu, B.~Zhang, R.~Xu, and J.~Li, ``Fairnas: Rethinking evaluation fairness
  of weight sharing neural architecture search,'' {\em arXiv preprint
  arXiv:1907.01845}, 2019.

\bibitem{Krizhevsky}
A.~Krizhevsky, V.~Nair, and G.~Hinton, ``Cifar-10 (canadian institute for
  advanced research),''

\bibitem{perez2017effectiveness}
L.~Perez and J.~Wang, ``The effectiveness of data augmentation in image
  classification using deep learning,'' {\em arXiv preprint arXiv:1712.04621},
  2017.

\end{thebibliography}
\bibliographystyle{ieeetr}}
%


%
%




\end{document}